\documentclass[10pt,twocolumn,letterpaper]{article}

\usepackage[pagenumbers]{cvpr} %

\usepackage{cuted} %

\usepackage{currfile} %

\usepackage{caption} %

\newcommand{\method}{\textsc{HapTwin}\xspace}

\newif\ifarxiv

\AtEndPreamble{
    \crefname{appendix}{Sec.}{Secs.}
    \Crefname{appendix}{Section}{Sections}
}

\arxivtrue

\definecolor{cvprblue}{rgb}{0.21,0.49,0.74}
\usepackage[pagebackref,breaklinks,colorlinks,allcolors=cvprblue]{hyperref}

\def\confName{CVPR}
\def\confYear{2027}

\makeatletter
\patchcmd{\@maketitle}{{\vspace*{24pt}}}{{\vspace*{4pt}}}{}{\PackageWarningNoLine{arxiv}{title gap patch failed}}
\makeatother

\newcommand{\projectpage}{\url{https://tt-232.github.io/HapTwin/}}

\title{
{\includegraphics[height=1.5\baselineskip]{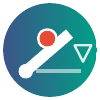}}\hspace{0.4em}%
\begin{tabular}{@{}c@{}}{\method}: Learning Articulated Object Dynamics\\from a Single Visuo-Haptic Hand Interaction\end{tabular}}
\author{
Tiantian Wei$^{1,3}$\,
Junsheng Ding$^{2,\dagger}$\,
Luca Collorone$^{1,3}$\,
\vspace{0pt} \\ 
Weihang Li$^{1,3}$\,
Yordanka Velikova$^{1,3}$\,
Benjamin Busam$^{1,3}$ \\
{\small $^{1}$\raisebox{-0.04em}{\includegraphics[height=0.70em]{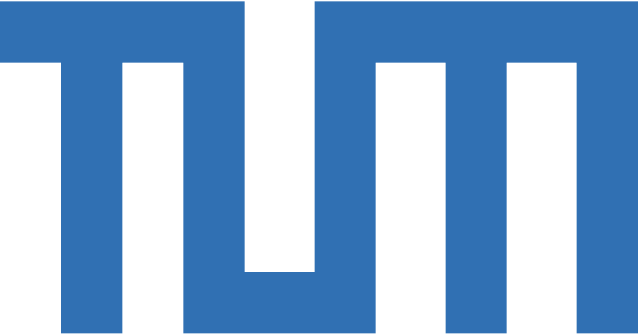}}\hspace{0.2em}Technical University of Munich\quad $^{2}$\raisebox{-0.02em}{\includegraphics[height=0.72em]{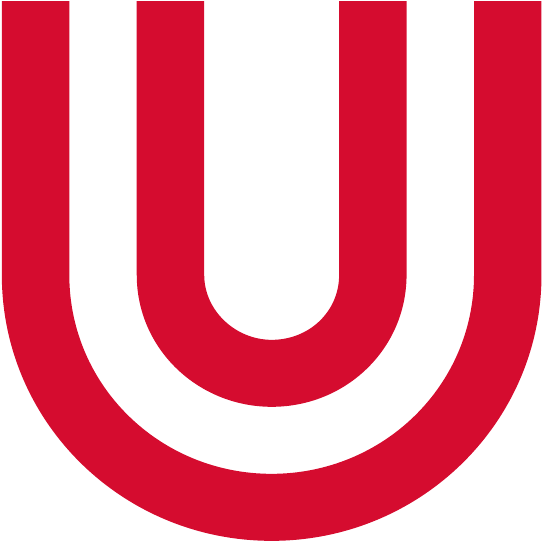}}\hspace{0.2em}University of Bremen\quad $^{3}$\raisebox{0.08em}{\includegraphics[height=0.50em]{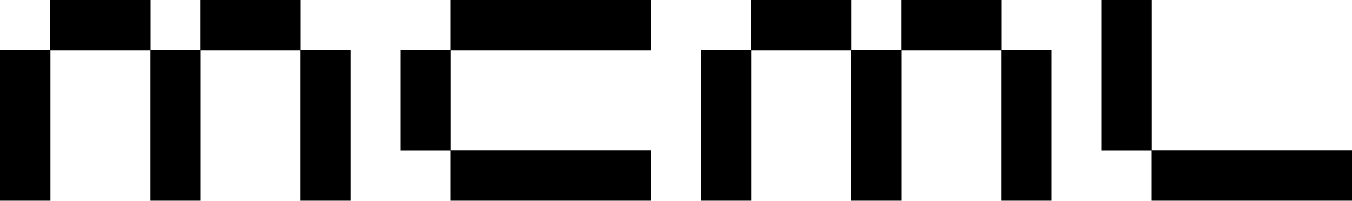}}\hspace{0.2em}MCML\qquad{\footnotesize $^{\dagger}$Corresponding author}} \\[-2pt]
{\footnotesize\texttt{\{tiantian.wei, b.busam\}@tum.de}\quad\texttt{jding@uni-bremen.de}}
\end{tabular}\endgraf\vspace*{-45pt}\begin{tabular}{c}
}

\begin{document}
\maketitle
\begin{strip}
\centering
\includegraphics[width=\linewidth]{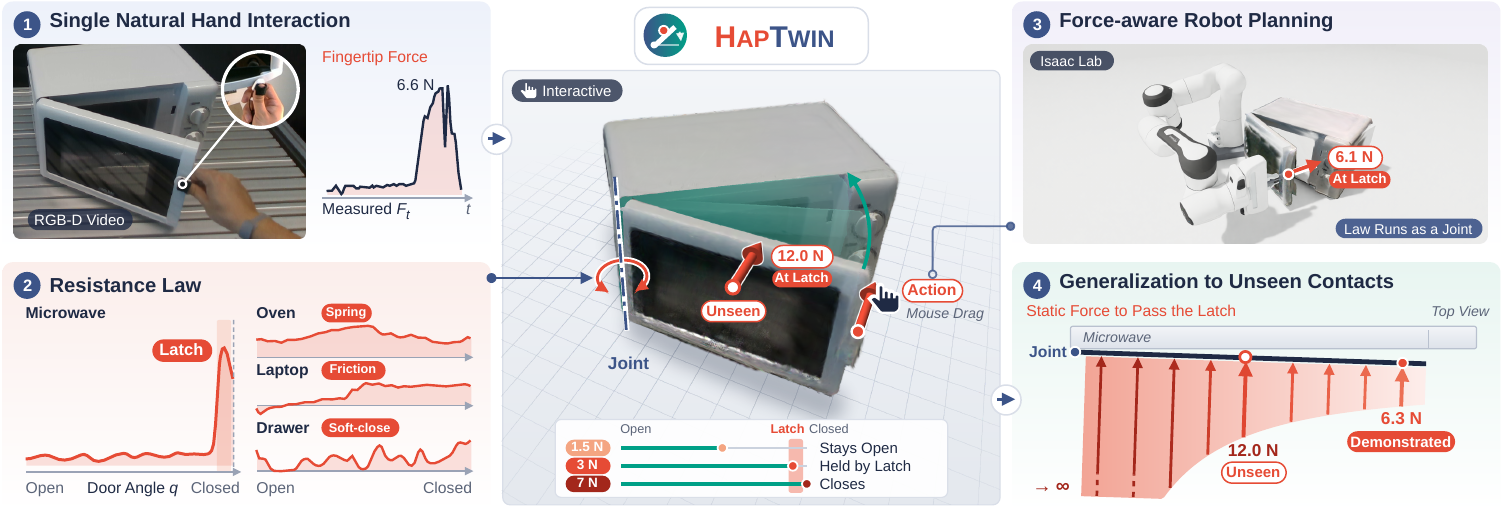}
\captionsetup{hypcap=false}
\captionof{figure}{\textbf{\method reconstructs a force-grounded twin from one natural hand interaction.} The twin (center) is interactive: a growing force applied by mouse drag leaves the door open, held by the latch, or closed. From RGB-D video and fingertip force~(1), it learns a free-form resistance law per object that captures internal mechanisms~(2), supports force-aware robot planning~(3), and generalizes zero-shot to unseen contact points~(4).}
\captionsetup{hypcap=true}
\label{fig:\currfilebase}%
\end{strip}

\begin{abstract}
Digital twins must reproduce not only how articulated parts move, but also how they resist applied forces. Our goal is simple: reconstruct articulated twins that resist. Existing reconstruction and generation approaches typically assign dynamics from defaults or learned priors, while force-based identification often relies on robotic actuation and designed excitation. We present \method, which reconstructs a force-grounded articulated twin from a single natural hand interaction recorded with RGB-D cameras and a fingertip force sensor. \method combines measured forces, surface-constrained contact, and per-frame poses 
to learn an object-specific joint resistance law. Differentiable rollouts jointly match observed poses, images, and contact forces across frames,
capturing resistance effects of friction, latches, and springs. Across nine interactions with four real objects, \method reduces generalized contact-force RMSE by $2.9\times$ on average over the strongest baseline. The resistance laws integrate seamlessly into standard robot simulators, and the twins generalize to a novel contact point, enabling force-aware planning beyond the demonstration. Project page: \projectpage

\vspace{-10pt}

\end{abstract}

\section{Introduction}
\label{sec:intro}

Digital twins of articulated objects serve as simulation assets for training, evaluating, and planning with embodied agents~\cite{xiang2020sapien,jin2026artvip,engelbracht2026forcetwin,xia2025drawer}, and as interactive content in AR/VR~\cite{jiang2022ditto,kiray2026promptvfx}.
For these applications, a twin must model not only how a part moves, but also the joint dynamics of how friction, inertia, and internal mechanisms such as latches, return springs, and dampers resist an applied force~\cite{engelbracht2026forcetwin}.
Recovering these dynamics from visual observation alone is underdetermined: internal mechanisms are hidden behind the 
appearance, and a door may open slowly under either a light push on a smooth hinge or a hard push on a stiff one, which only the contact force can distinguish.

Existing methods for building articulated twins face a trade-off between scalable capture and physical grounding.
Visual reconstruction and generation methods~\cite{xia2025drawer,le2025articulateanything,he2026neoworldpro,cai2026artisan,delitzas2026funrec,jin2026artvip,gao2026joda} recover geometry and kinematics from easily captured images or videos, and assign joint dynamics from simulator defaults, manual tuning, or vision-language priors.
They yield dynamics plausible for the object's appearance but do not identify the object-specific resistance from measured interaction forces.
Conversely, system identification methods~\cite{martin2022coupled,he2026rigpi,engelbracht2026forcetwin,baum2026admittance} measure motion and force precisely to recover
joint dynamics that generalize to new actions in standard simulators~\cite{engelbracht2026forcetwin,he2026rigpi}, but rely on designed excitation through robotic actuation or instrumented tools.
For deformable objects, PhysTwin~\cite{jiang2025phystwin} resolves this trade-off by modeling the hand in a natural interaction as the driving action, and the learned twin predicts the response to unseen actions.
This raises a natural question: can we recover force-grounded joint dynamics of an articulated object directly from a single natural hand interaction?

The challenge is that identifying joint dynamics requires the part motion, the contact point, and the contact force, which a hand--object interaction video observes with different uncertainty.
The per-frame part pose is constrained by observations across the visible surface, but fitting the dynamics requires its velocity and acceleration, and differentiating poses across frames amplifies the noise in them.
Errors in the contact point or contact force corrupt the applied torque, yet the contact point is only approximate due to the small and often occluded fingertips, and the contact force cannot be observed visually.

Our key idea is to reduce the uncertainty in each quantity required by the joint dynamics through the choice of observation and formulation.
We present \method (\cref{fig:teaser}), to our knowledge the first method to reconstruct a force-grounded articulated object twin from a single natural visuo-haptic hand interaction.
We measure the contact force with a fingertip force sensor and constrain the contact point to the reconstructed surface of the part.
Instead of estimating velocity and acceleration from the poses, we learn a free-form resistance law over the joint coordinate in differentiable simulation, jointly supervised by the per-frame images, poses, and measured forces.
The learned law captures resistance effects of latches, return springs, and friction hinges, and the twin predicts both the part motion and the contact force from new actions of a hand or a robot.

In summary, our contributions are as follows:
\begin{itemize}
  \item \textbf{Force-grounded reconstruction.} We recover an articulated digital twin with its joint resistance from a single natural hand interaction.
  \item \textbf{Resistance learning.} We learn a free-form joint resistance law in differentiable simulation by combining measured contact forces with a surface-constrained contact point.
  \item \textbf{Predictive capability.} The twins reproduce the observed part motion and the measured contact force, with $2.9\times$ lower force error than the strongest baseline, transfer to a standard robot simulator, and generalize zero-shot to novel contact points.
\end{itemize}

\section{Related Work}
\label{sec:related}

\noindent\textbf{Articulated object reconstruction and generation.}
Reconstruction methods recover part-level geometry, appearance, and kinematics of real objects from multi-state images~\cite{liu2023paris,liu2025artgs,li2026gear} or human interaction videos~\cite{liu2025videoartgs,peng2026itaco,dai2026freeartgs,werby2025artipoint,kerr2024rsrd,wang2026arthoi}, while generation methods synthesize articulated assets at scale~\cite{lei2023nap,liu2024cage,chen2024urdformer,qiu2025articulateanymesh,le2025articulateanything}.
Notably, iTACO~\cite{peng2026itaco} masks out the hand in the interaction video, as kinematics can be recovered without the driving force.
To support simulation~\cite{mandi2025real2code,xia2025drawer,delitzas2026funrec,zhang2026trackarticulateact,jin2026artvip}, recent works in both lines add physical parameters, ranging from default~\cite{jiang2022ditto} or manually tuned values~\cite{jin2026artvip} to per-object values inferred from appearance or text via foundation-model priors~\cite{delitzas2026funrec,xu2025gaussianproperty,cao2025physx3d,cao2026physxanything,yang2026physforge,zhang2026simart,zhou2026articraft,he2026neoworldpro}.
Going further, Artisan~\cite{cai2026artisan} estimates part mass, joint friction, and damping to generate physically plausible assets, while JODA~\cite{gao2026joda} composes position-dependent conservative force, friction, and damping from effects proposed by a vision-language model.
Such inference needs no physical interaction and scales to assets without real-world counterparts.
However, the resistance of a specific real object comes from hidden internal mechanisms and is revealed by its response to an applied force.
We complement both lines by modeling the hand as the driving action and learning an object-specific, state-dependent resistance law from the measured contact force.

\noindent\textbf{Physics from language and video.}
While open-vocabulary prompts can specify the physics of a 3D Gaussian scene in real time~\cite{collorone2025phystalk}, physics-informed reconstruction learns the material parameters of deformable objects from video~\cite{li2023pacnerf,zhong2024springgaus,zhang2024physdreamer}.
PhysTwin~\cite{jiang2025phystwin} goes further and models the visually tracked hand in a natural human interaction as the driving action, so the action-conditioned twin predicts its response to unseen actions and supports robot planning and policy evaluation~\cite{zhang2026real2simeval}.
These twins couple the object to the tracked hand by springs and are learned without measuring the contact force, as the visible deformation of the soft object provides additional observable constraints on the dynamics.
A rigid part, however, does not deform, and estimating force from vision without a sensor~\cite{grady2022pressurevision,zhao2025egopressure,ilic2026egophi,zhang2026egotac} provides only a learned visual prior rather than an object-specific measurement.
Inheriting this action-conditioned view, we extend it to force-grounded articulated twins by measuring the contact force with a fingertip force sensor and constraining the contact point to the reconstructed rigid surface.

\noindent\textbf{Instrumented interaction and identification.}
Robotics has long characterized how articulated objects resist manipulation~\cite{jain2010complex,endres2013doors,martin2022coupled}, and recent systems identify joint dynamics from measured contact forces~\cite{engelbracht2026forcetwin,he2026rigpi,baum2026admittance}, which generalize zero-shot to new actions in standard simulators~\cite{engelbracht2026forcetwin,he2026rigpi}.
These systems measure the applied wrench with a force--torque sensor rigidly mounted on a robot arm, on a handheld gripper, or at the object handle, and track the motion with proprioception, SLAM, or motion capture.
Wrench-based methods~\cite{engelbracht2026forcetwin,baum2026admittance} fit dynamics using differentiated pose trajectories, whereas RigPI~\cite{he2026rigpi} identifies constant joint friction through differentiable simulation.
Tactile gloves and force-sensing grippers record the force of natural human interactions~\cite{li2024learning,engelbracht2025hoi,zhang2026glove2hand}, but these works do not identify object-specific joint resistance.
We combine a lightweight wearable fingertip force sensor with surface-constrained contact to learn a state-dependent joint resistance law from a single natural interaction through rollouts supervised by observed motion and measured force,
without differentiating observed poses.

\section{Method}
\label{sec:method}

\begin{figure*}[t]
  \centering
  \includegraphics[width=\linewidth]{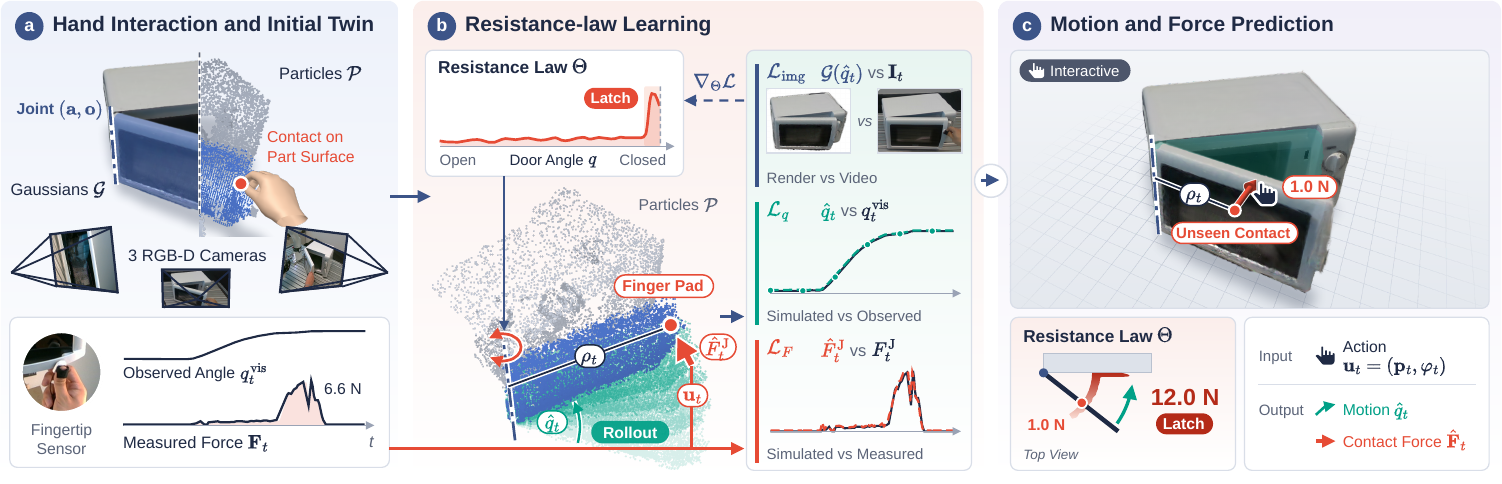}
  \vspace{-20pt}
  \caption{\textbf{Overview.} (a) From one natural hand interaction, recorded by RGB-D cameras and a fingertip force sensor, we reconstruct two-part Gaussians $\mathcal{G}$, the joint $(\mathbf{a}, \mathbf{o})$, the per-frame joint coordinate $q^{\text{vis}}_t$, and the simulation particles $\mathcal{P}$. (b) The measured force drives a finger pad on the part surface, and we learn the resistance law $\Theta$ through differentiable simulation from the images, $q^{\text{vis}}_t$, and the force. (c) The learned twin predicts the part motion $\hat{q}_t$ and the contact force $\hat{\mathbf{F}}_t$ for new finger actions $\mathbf{u}_t$, also at an unseen contact point, without refitting.}
  \label{fig:pipeline}
  \vspace{-10pt}

\end{figure*}

\subsection{Problem Formulation}
\label{sec:problem}
We consider an articulated object with one rigid moving part, such as a door, a lid, or a drawer, attached to a static body by a single-degree-of-freedom (1-DoF) joint, either revolute or prismatic.
The joint has unit axis $\mathbf{a} \in \mathbb{R}^3$ through the point $\mathbf{o} \in \mathbb{R}^3$, and its coordinate $q \in \mathbb{R}$, measured from the pose of the part in the first frame, is an angle (rad) for a revolute joint and a displacement (m) for a prismatic one.

A person operates the part once by hand as in everyday use, and calibrated RGB-D cameras and a fingertip force sensor record this interaction as $M$ time-aligned frames $\mathcal{D} = \{(\mathbf{I}_t, \mathbf{D}_t, \mathbf{F}_t)\}_{t=1}^{M}$, with the color images $\mathbf{I}_t$ and the depth maps $\mathbf{D}_t$ of the cameras and the measured contact force $\mathbf{F}_t \in \mathbb{R}^3$.

From $\mathcal{D}$ we reconstruct the twin $\mathcal{T} = (\mathcal{G}, (\mathbf{a}, \mathbf{o}), \mathcal{P}, \Theta)$, with the 3D Gaussians $\mathcal{G}$ of the body and the part, the joint $(\mathbf{a}, \mathbf{o})$, the simulation particles $\mathcal{P}$, and the resistance law $\Theta$, which gives the generalized force that the joint and its internal mechanism exert on the part as a function of the joint state.
Given an action $\mathbf{u}_t = (\mathbf{p}_t, \varphi_t) \in \mathbb{R}^3 \times \mathbb{R}$, with a contact point $\mathbf{p}_t$ on the part and a finger position $\varphi_t$ along the joint coordinate $q$, the twin predicts the part motion $\hat{q}_t$ and the contact force $\hat{\mathbf{F}}_t \in \mathbb{R}^3$ and renders the object (\cref{fig:pipeline}).

\subsection{Geometric and Kinematic Reconstruction}

\label{sec:recon}

From $\{(\mathbf{I}_t, \mathbf{D}_t)\}_{t=1}^{M}$, we reconstruct the initial twin $(\mathcal{G}, (\mathbf{a}, \mathbf{o}), \mathcal{P})$ and the per-frame joint coordinate $q^{\text{vis}}_t$ (\cref{fig:pipeline}a, more details in \cref{sec:supp_recon}).

\noindent\textbf{Segmentation and joint initialization.}
We segment the object and the hand in every view~\cite{ren2024groundedsam,ravi2025sam2} and track 3D points of the object~\cite{karaev2024cotracker3}, lifted with depth. The tracks separate the moving part from the static body and give an initial joint $(\mathbf{a}, \mathbf{o})$. A MANO hand~\cite{romero2017mano}, fitted to the WiLoR~\cite{potamias2025wilor} keypoints of all cameras locates the contact point $\mathbf{p}_t$ (\cref{sec:dynamics}).

\noindent\textbf{Gaussian reconstruction.}
The Gaussians $\mathcal{G}$ start from generated meshes~\cite{li2026mvsam3d,sam3d2025} and the observed depth.
We train them together with $(\mathbf{a}, \mathbf{o})$ and $q^{\text{vis}}_t$ on all cameras and frames, with the part posed by the joint at $q^{\text{vis}}_t$.
The image loss $\mathcal{L}_{\text{rec}}$ combines a color error $\mathcal{L}_{\text{color}}$ outside the hand mask, a silhouette error $\mathcal{L}_{\text{sil}}$ against the object mask, and a Huber depth error $\mathcal{L}_{\text{depth}}$,
\vspace{-5pt}
\begin{equation}
  \mathcal{L}_{\text{rec}} = \mathcal{L}_{\text{color}} + \lambda_{\text{sil}}\,\mathcal{L}_{\text{sil}} + \lambda_{\text{depth}}\,\mathcal{L}_{\text{depth}}.
  \label{eq:img}
  \vspace{-5pt}
\end{equation}
We use novel views restored by GeoQuery~\cite{cao2026geoquery} and silhouettes of the generated meshes to regularize the unobserved regions.

\noindent\textbf{Simulation particles.}
The simulation particles $\mathcal{P}$ come from the tracked 3D points, completed from the Gaussians where the tracks miss the surface.
Thus $\mathcal{G}$, $(\mathbf{a}, \mathbf{o})$, and $\mathcal{P}$ come from one unified reconstruction, providing an accurate surface to anchor the fingertip contact.
Because the part is rigid, its particles and Gaussians move together with the joint, and the state of the object can be described by $(q, \dot{q})$. The part particles give the moment of inertia $I$ of the part and its gravity torque $\tau_{\text{grav}}(q) = A\cos q + B\sin q$ with constant $A$ and $B$.

\subsection{Resistance-Law Learning in Simulation}
\label{sec:dynamics}
\label{sec:learning}

We learn the resistance law $\Theta$ in a differentiable simulation (\cref{fig:pipeline}b), where the measured force $\mathbf{F}_t$ drives a finger pad on the reconstructed part and the video and the sensor supervise its motion and force
(More details in \cref{sec:supp_learning}).

\noindent\textbf{Joint equation and resistance law.}
We define $\tau_c$ as the generalized contact force that the finger applies to the part, and model the force required to realize the joint motion as
\vspace{-5pt}
\begin{equation}
  \tau_c =
  \underbrace{I\,\ddot{q} - \tau_{\text{grav}}(q)}_{\text{rigid-body terms}}
  + \underbrace{c\,\dot{q} + f(q)\,\operatorname{sgn}(\dot{q}) - g(q)}_{\text{resistance law}}.
  \label{eq:eom}
\end{equation}
The resistance law consists of a viscous damping $c \ge 0$, a Coulomb friction $f(q) \ge 0$, and the torque $g(q) = -\mathrm{d}V/\mathrm{d}q$ of a potential $V(q)$.
The potential covers every conservative torque beyond the gravity torque, such as a spring, a counterweight, or a detent.
For a prismatic joint, the same equation holds with $I$ the mass of the part, $\tau_{\text{grav}} = A$ the weight along the axis, and generalized forces in N.

With the reconstructed geometry, the joint, and the assigned mass properties fixed, we optimize the resistance law.
The hidden internal mechanisms make the resistance vary with $q$, so we parameterize $V(q)$ and $f(q)$ in free form as two MLPs, with a softplus on the output of $f(q)$, and learn their weights and $c$, which parameterize $\Theta$.
The law can thus store and dissipate energy without generating it.
We initialize the law as a standard simulator joint with constant friction $f_{0}$, damping $c$, and $g(q) = 0$ so that the configuration-dependent resistance is learned from the data.

\noindent\textbf{Fingertip contact.}
Video alone determines neither the contact force nor the precise contact between the fingertip and the part, so driving a rigid part with the tracked fingertip would turn its position error directly into a force error.
We therefore anchor the finger as a compliant pad on the reconstructed part surface, driven by the measured force $\mathbf{F}_t$. On this surface, the fitted hand locates the contact point $\mathbf{p}_t$ tangentially, so its error only slightly changes the lever arm.

The joint lets $\mathbf{p}_t$ move along $\mathbf{J}_t = \partial \mathbf{p}_t / \partial q$ with the lever arm $\rho_t = \lVert \mathbf{J}_t \rVert$.
The pad is a unilateral spring-damper that follows the finger position $\varphi_t$ and pushes on the part only when compressed, and its normal force $F_n$ acts on the joint through the lever arm $\rho_t$ as the generalized contact force $\tau_c$ of \cref{eq:eom},
\begin{equation}
  F_n = \big[k_n\,\delta_t + c_n\,\dot{\delta}_t\big]_{+},
  \qquad
  \tau_c = s_t\,\rho_t\,F_n,
  \label{eq:contact}
\end{equation}
where $\delta_t \approx s_t\,\rho_t\,(\varphi_t - q)$ is the compression of the pad, $k_n$ (N/m) and $c_n$ (N\,s/m) are its stiffness and damping, and $s_t \in \{\pm 1\}$ is the direction in which the hand moves the part.
The simulated contact force on the part is $\hat{\mathbf{F}}_t = s_t\,F_n\,\mathbf{J}_t / \rho_t$, the pad force along the direction in which the contact point moves.
On the contact frames $\mathcal{C}$, where the sensed finger drives the part, we place the finger so that the pad's elastic contribution matches the measured force at the smoothed per-frame joint coordinate $\bar{q}^{\text{vis}}_t$.
With the component $F^{\text{J}}_t = s_t\,\mathbf{F}_t^{\top}\mathbf{J}_t / \rho_t$ of the measured force along the motion of the contact point, the finger leads $\bar{q}^{\text{vis}}_t$ by the compression that this force requires,
\vspace{-5pt}
\begin{equation}
  \varphi_t = \bar{q}^{\text{vis}}_t + s_t\,\frac{F^{\text{J}}_t}{k_n\,\rho_t}.
  \label{eq:pad}
\end{equation}
An error of the law then moves the simulated part away from the observed one and changes the pad force.

\noindent\textbf{Differentiable simulation.}
We learn the law through rollouts of the twin, the same way the twin is used later.
From every training frame $t_0$, we start a rollout window at the smoothed observed state $(\bar{q}^{\text{vis}}_{t_0}, \dot{\bar{q}}^{\text{vis}}_{t_0})$ and integrate \cref{eq:eom} for $K = 32$ frames under the action $\mathbf{u}_t = (\mathbf{p}_t, \varphi_t)$.
Each window yields the simulated coordinate $\hat{q}_{t}$ and the contact force $\hat{\mathbf{F}}_{t}$ of its frames, which we compare with the per-frame joint coordinate $q^{\text{vis}}_t$, the images $(\mathbf{I}_t, \mathbf{D}_t)$, and the measured force $\mathbf{F}_t$, whose component $F^{\text{J}}_t$ is compared with its simulated counterpart $\hat{F}^{\text{J}}_t = F_n$,
\vspace{-5pt}
\begin{align}
  \mathcal{L}_{q} &= \big\langle (\hat{q}_{t} - q^{\text{vis}}_{t})^{2} \big\rangle\,,
  \qquad
  \mathcal{L}_{\text{img}} = \frac{\big\langle \mathcal{L}_{\text{rec}}(\hat{q}_{t}) \big\rangle}{\Delta_{\text{img}}}\,,
  \label{eq:lq}\\
  \mathcal{L}_{F} &= \big\langle \big(\log(\hat{F}^{\text{J}}_{t} + \epsilon) - \log(F^{\text{J}}_{t} + \epsilon)\big)^{2} \big\rangle_{t \in \mathcal{C}}\,,
  \label{eq:lf}
\end{align}
where $\langle\cdot\rangle$ is the mean over the frames of all windows, restricted to $\mathcal{C}$ for the force, and $\Delta_{\text{img}}$ is the increase of $\mathcal{L}_{\text{rec}}$ (\cref{eq:img}) when the part is $1^\circ$ off its per-frame joint coordinate.
Both forces are non-negative on $\mathcal{C}$, and the logarithm with a small offset $\epsilon > 0$ penalizes relative errors, so a light action counts as much as the peak of a latch.

An action in one direction observes only the combined resistance of friction and potential, yet its decomposition determines whether a released part stays in place or moves back.
The rest frames $\mathcal{R}$, where the hand exerts no force and the part stays still, provide additional constraints on this decomposition. We use them to regularize the net conservative torque $N(q) = g(q) + \tau_{\text{grav}}(q)$ toward zero, selecting equilibria that do not rely on static friction.
$\mathcal{L}_{\text{rest}}$ penalizes $(N(q^{\text{vis}}_t)/\tau_0)^2$ on these frames, except a torque into a stop that holds the part, with a torque scale $\tau_0$.
Where the part never rests, a weak prior $\mathcal{L}_{V} = \langle (g(q)/\tau_0)^2 \rangle_{q}$ over the range of motion selects the law with the smallest conservative torque, so a potential enters the law only where the force, the motion, or the rest frames require it.
The total loss is
\begin{equation}
  \mathcal{L} = \mathcal{L}_{\text{img}} + \lambda_{q}\,\mathcal{L}_{q} + \lambda_{F}\,\mathcal{L}_{F} + \mathcal{L}_{\text{rest}} + \lambda_{V}\,\mathcal{L}_{V}.
  \label{eq:total}
\end{equation}

\subsection{Force-Grounded Simulation}
\label{sec:sim}

With the learned law, our particle simulator integrates \cref{eq:eom} under the pad contact of an action $\mathbf{u}_t$ (\cref{eq:contact}), predicting the part motion $\hat{q}_t$ and the contact force $\hat{\mathbf{F}}_t$ and rendering the twin at $\hat{q}_t$ (\cref{fig:pipeline}c).
Because the resistance law is parameterized in standard joint-space torque, friction, and damping, it executes directly in standard articulated simulators (\cref{sec:exp_sim2sim}). Moreover, a new contact point changes only the lever arm $\rho_t$ and leaves the law of the joint unchanged, so the twin generalizes zero-shot to actions at unseen contact points (\cref{sec:exp_cosim}).

\section{Experiments}
\label{sec:exp}

\subsection{Experiment Settings}
\label{sec:exp_settings}

\begin{figure*}[t]
  \centering
  \includegraphics[width=\linewidth]{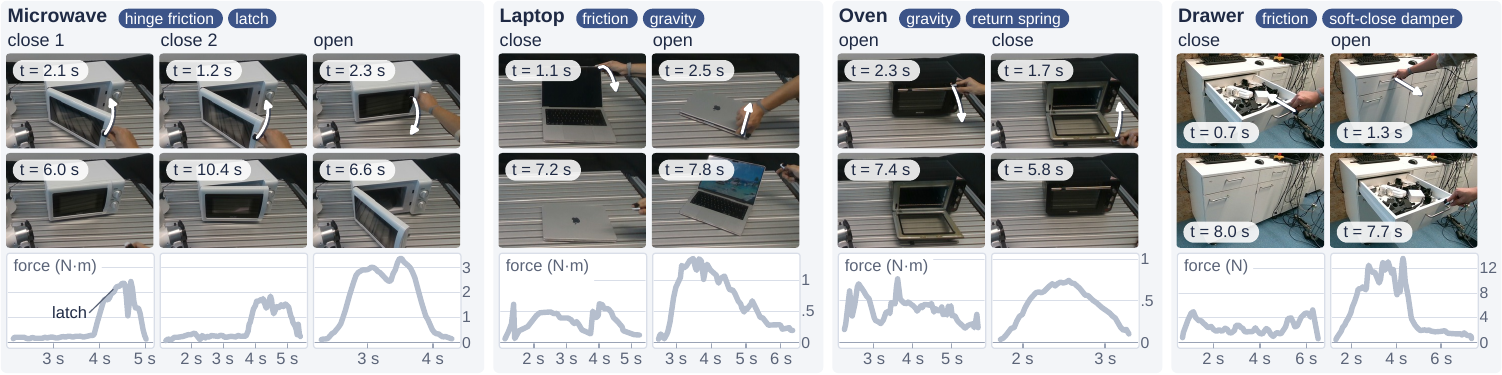}

  \caption{\textbf{Dataset overview.} For each of the nine takes in \cref{tab:main}, we show the frame at contact onset with the direction of the finger motion (white arrow), the last frame, and the measured generalized contact force. Takes of the same object share the force axis.}
  \label{fig:dataset}
  \vspace{-5pt}
\end{figure*}

\noindent\textbf{Dataset.}
To cover diverse internal mechanisms, we record a person opening and closing four everyday objects across revolute and prismatic joints: a microwave with a latch, a laptop with a friction hinge, an oven with a return spring, and a drawer with a soft-close damper.
Three Intel RealSense cameras record RGB-D video at 30\,fps (one for the drawer), and a PaXini PX-6AX GEN3 on one fingertip records the 3D contact force at 30\,Hz.
\Cref{fig:dataset} shows the nine takes, each a single natural hand interaction of 5.8 to 10.5\,s.

\noindent\textbf{Metrics.}
We evaluate the open-loop rollout of each method in appearance, motion, and contact force.
Appearance is measured by PSNR, and motion by the average articulation error (AAE)~\cite{fan2023arctic} against the per-frame joint coordinate $q^{\text{vis}}_t$ and the Chamfer distance of the moving part (CD-m)~\cite{li2026gear,dai2026freeartgs}.
For the contact force, we compare the measured and predicted generalized force on the joint, $\tau_t = \rho_t F^{\text{J}}_t$ and $\hat{\tau}_t = \rho_t \hat{F}^{\text{J}}_t$, both along the direction in which the hand moves the part, and score all frames of $\mathcal{C}$ with the RMSE, which measures the magnitude of the force error, and with the Pearson correlation $r$, which measures how well the prediction follows the shape of the measured force (\cref{sec:supp_exp_recon}).

\subsection{Twin Reconstruction from One Interaction}
\label{sec:exp_recon}

\begin{table*}[t]
  \caption{\textbf{Quantitative force results on twin reconstruction.} We compare the contact force of each open-loop rollout with its training take, for the baselines (A)--(D), adapted from~\cite{zhang2026egotac,cai2026artisan,engelbracht2026forcetwin,jiang2025phystwin}, the ablations (E) and (F), and \method. RMSE in N$\cdot$m (N for the drawer). $r$ is the Pearson correlation with the measured force. The last row is the mean over the takes, with each RMSE as a multiple of ours.}
  \vspace{-5pt}

  \centering
  \footnotesize
  \setlength{\tabcolsep}{1.8pt}
  \renewcommand{\arraystretch}{0.92}
  \newcommand{\meanband}{\noalign{\vskip-\belowrulesep
    \begingroup\color{black!9}\hrule height\dimexpr\belowrulesep+\arraystretch\ht\strutbox+\arraystretch\dp\strutbox+\aboverulesep\relax\endgroup
    \vskip-\dimexpr\arraystretch\ht\strutbox+\arraystretch\dp\strutbox+\aboverulesep\relax}}
  \begin{tabular}{ll cc cc cc cc @{\hskip 7pt} cc cc @{\hskip 7pt} cc}
    \toprule
    & & \multicolumn{8}{c}{Baselines} & \multicolumn{4}{c}{Ablations} & \multicolumn{2}{c}{Ours} \\
    \cmidrule(lr){3-10}\cmidrule(lr){11-14}\cmidrule(lr){15-16}
    & & \multicolumn{2}{c}{(A) \texttt{VisionForce}} & \multicolumn{2}{c}{(B) \texttt{AssetGen}} & \multicolumn{2}{c}{(C) \texttt{SysID}} & \multicolumn{2}{c}{(D) \texttt{VideoTwin}} & \multicolumn{2}{c}{(E) \texttt{MANOContact}} & \multicolumn{2}{c}{(F) \texttt{ConstantLaw}} & \multicolumn{2}{c}{\method} \\
    Object & Take & RMSE$\downarrow$ & $r\uparrow$ & RMSE$\downarrow$ & $r\uparrow$ & RMSE$\downarrow$ & $r\uparrow$ & RMSE$\downarrow$ & $r\uparrow$ & RMSE$\downarrow$ & $r\uparrow$ & RMSE$\downarrow$ & $r\uparrow$ & RMSE$\downarrow$ & $r\uparrow$ \\
    \midrule
    Microwave & close 1 & 0.643 & 0.59 & 0.580 & 0.74 & 0.355 & 0.89 & 0.927 & 0.05 & 0.420 & 0.83 & 0.574 & 0.71 & \textbf{0.060} & \textbf{1.00} \\
     & close 2 & 0.646 & $-$0.02 & 0.797 & $-$0.32 & 0.120 & 0.98 & 0.635 & 0.62 & 0.331 & 0.90 & 0.760 & $-$0.34 & \textbf{0.080} & \textbf{0.99} \\
     & open & 1.767 & 0.55 & 2.023 & 0.11 & 0.566 & 0.94 & 2.107 & $-$0.13 & 1.687 & 0.38 & 1.981 & $-$0.08 & \textbf{0.286} & \textbf{0.97} \\
    \midrule
    Laptop & close & 0.374 & $-$0.11 & 0.208 & 0.38 & 0.156 & 0.62 & 0.352 & $-$0.18 & 0.080 & 0.85 & 0.637 & 0.35 & \textbf{0.057} & \textbf{0.93} \\
     & open & 0.401 & 0.51 & 0.274 & 0.88 & 0.148 & 0.94 & 0.936 & $-$0.19 & 0.259 & 0.79 & 0.190 & 0.93 & \textbf{0.073} & \textbf{0.99} \\
    \midrule
    Oven & open & 0.474 & 0.42 & 0.401 & $-$0.41 & 0.243 & 0.31 & 0.413 & $-$0.07 & 0.844 & $-$0.51 & 0.400 & $-$0.41 & \textbf{0.086} & \textbf{0.89} \\
     & close & 0.353 & $-$0.45 & 0.605 & 0.63 & 0.116 & 0.88 & 1.362 & 0.33 & 0.212 & 0.76 & 0.758 & 0.59 & \textbf{0.043} & \textbf{0.98} \\
    \midrule
    Drawer & close & 2.287 & 0.13 & 2.145 & 0.46 & 0.811 & 0.75 & 3.165 & $-$0.61 & 1.319 & 0.61 & 2.066 & 0.39 & \textbf{0.227} & \textbf{0.98} \\
     & open & 6.519 & 0.03 & 6.828 & $-$0.26 & 4.197 & 0.61 & 12.589 & 0.14 & 5.405 & 0.41 & 6.828 & $-$0.26 & \textbf{1.627} & \textbf{0.94} \\
    \midrule
    \meanband
    \multicolumn{2}{l}{Mean} & 7.2$\times$ & 0.18 & 7.4$\times$ & 0.25 & 2.9$\times$ & 0.77 & 12.0$\times$ & 0.00 & 5.1$\times$ & 0.56 & 8.4$\times$ & 0.21 & \textbf{1.0$\times$} & \textbf{0.96} \\
    \bottomrule
  \end{tabular}
  \label{tab:main}
  \vspace{-10pt}

\end{table*}

\begin{figure}[t]
  \centering
  \includegraphics[width=\linewidth]{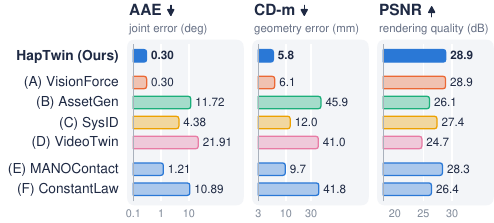}
  \vspace{-15pt}
  \caption{\textbf{Motion and appearance on twin reconstruction.} Means over the takes of \cref{tab:main} for the open-loop rollout of each method. AAE is averaged over the seven revolute takes, and AAE and CD-m are on a logarithmic axis. Per-take values are in the supplementary (\cref{sec:supp_exp_recon}).}
  \label{fig:motion}
  \vspace{-15pt}

\end{figure}

\begin{figure*}[t]
  \centering
  \includegraphics[width=\linewidth]{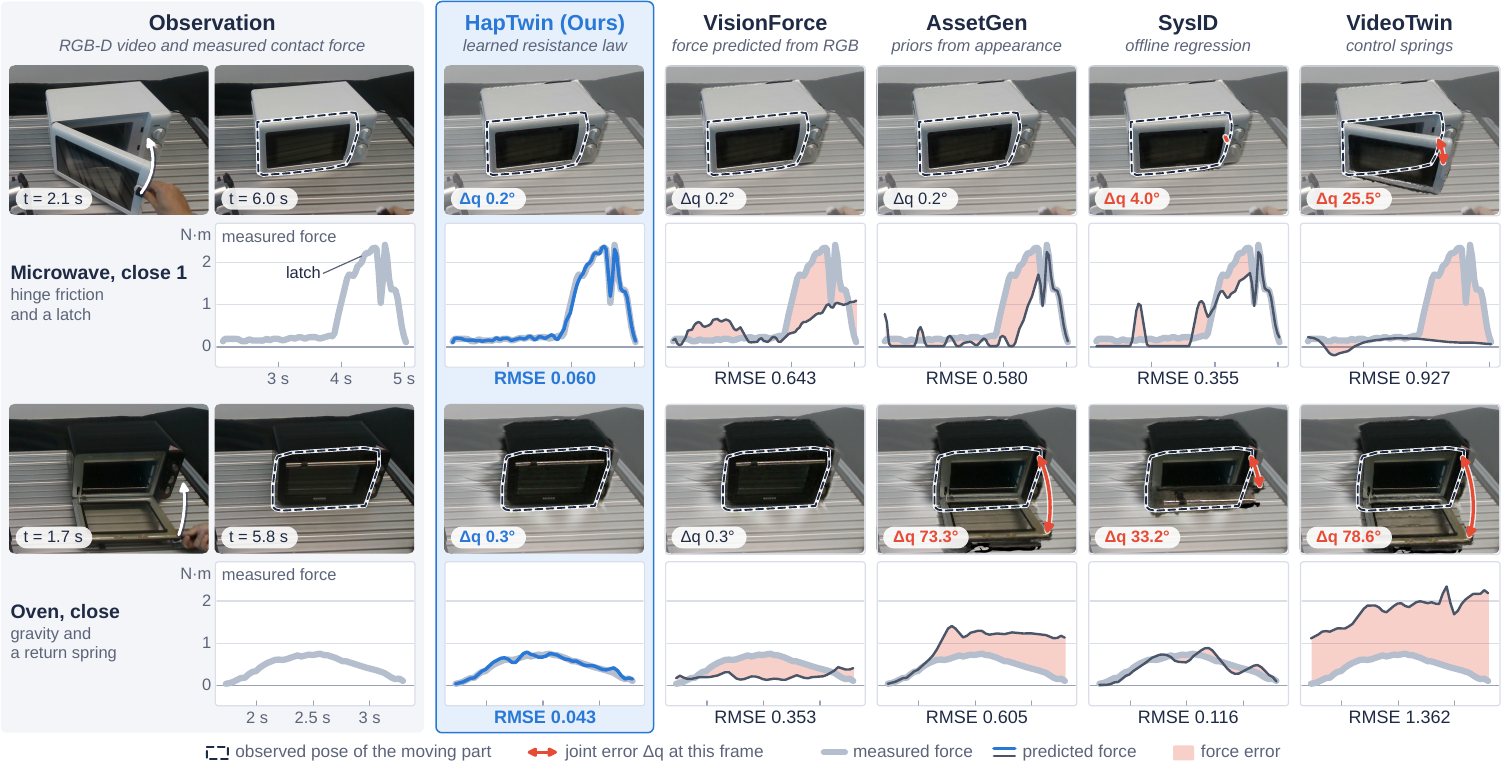}

  \caption{\textbf{Qualitative results on twin reconstruction.} We visualize the open-loop rollout at the end of the take and the generalized contact force of each method on microwave close~1 (top) and oven close (bottom). Compared with the baselines, only our predicted force closely follows the measured force (gray), including the latch peak (top) and the return-spring torque (bottom).}
  \label{fig:replay}
  \vspace{-10pt}
\end{figure*}

We first evaluate the reconstruction fidelity of a twin learned from a single interaction, in appearance and motion and, most importantly, in contact force.

\noindent\textbf{Setup.} Each of the nine twins is initialized at contact onset of its training take (\cref{fig:dataset}), driven in an open-loop rollout by the recorded 3D finger action, and compared with the recordings of that take.

\noindent\textbf{Baselines and ablations.}
To the best of our knowledge, no existing method learns the joint dynamics of an articulated object from a natural hand interaction and predicts both the motion and the contact force, so none applies to our setting directly.
We therefore adapt four representative lines of work to our setting as baselines and further ablate two designs of our own method.
All variants share our joint, particles, and Gaussians, so each difference in the results traces back to the replaced design (details in \cref{sec:supp_exp_recon}).
\begin{itemize}
  \item \emph{Can visual or foundation-model priors replace \textbf{direct force measurement}?}
  (A) \texttt{VisionForce}, adapted from EgoTac~\cite{zhang2026egotac}, replaces the sensor magnitude with the fingertip force predicted zero-shot from RGB video.
  (B) \texttt{AssetGen}, a system-level baseline adapted from Artisan~\cite{cai2026artisan}, replaces our learned resistance law with the constant mass, friction, and damping that its agents assign from appearance.

  \item \emph{Can offline regression on differentiated poses replace \textbf{learning in the simulation loop}?}
  (C) \texttt{SysID}, adapted from ForceTwin~\cite{engelbracht2026forcetwin}, replaces our in-the-loop rollout optimization with frame-by-frame regression of the measured torque on numerically differentiated poses.

  \item \emph{Can coupling to the tracked hand replace \textbf{surface-constrained contact}?}
  (D) \texttt{VideoTwin}, a system-level baseline adapted from PhysTwin~\cite{jiang2025phystwin}, replaces our pad contact, resistance law, and training with the control springs attached to the tracked 3D hand, the constant physical parameters, and the two-stage optimization of PhysTwin, and adds our measured force loss.
  To further isolate the sensitivity of rigid-body dynamics to fingertip tracking errors, the ablation (E) \texttt{MANOContact} places the contact point directly at the fitted MANO fingertip.

  \item \emph{Can constant joint parameters, instead of a \textbf{free-form resistance law}, capture the configuration-dependent resistance of internal mechanisms?}
  The ablation (F) \texttt{ConstantLaw} uses a constant friction $f$, a linear spring of constant stiffness $k$, and a constant damping $c$ in \cref{eq:eom}.
\end{itemize}

\noindent\textbf{Results.}
In motion and appearance, \method is on par with the best baseline, \texttt{VisionForce} (\cref{fig:motion}): both reach a mean AAE of $0.30^\circ$ and a PSNR of 28.9\,dB, while the other variants deviate by $1.2^\circ$ to $21.9^\circ$.
The contact force is where the methods differ (\cref{tab:main}): \method achieves the lowest force RMSE, on average $2.9\times$ lower than the strongest baseline \texttt{SysID} ($1.5\times$--$5.9\times$ per take), and the highest correlation ($r = 0.89$--$1.00$) on all nine takes.
Without mechanism-specific templates, the learned law captures internal mechanics, e.g., the latch peak at the end of \textit{microwave close~1} and the return-spring torque throughout \textit{oven close} (\cref{fig:replay}, more in the supplementary).
Since internal mechanisms are not directly observable from appearance, visual and foundation-model priors cannot replace direct force measurement: \texttt{VisionForce} tracks the motion but incurs up to $11\times$ higher force RMSE, while \texttt{AssetGen} models constant friction, damping, and mass but misses these angle-dependent mechanics.
Despite also using measured forces, \texttt{SysID} performs offline regression on differentiated poses rather than learning in the simulation loop, fitting amplified velocity and acceleration noise and causing rollout drift ($r = 0.31$ on \textit{oven open}).
Because rigid parts cannot deform to absorb hand-tracking noise, the hand--object spring coupling of \texttt{VideoTwin} turns tracking errors into spring forces, causing negative force correlations on five takes and severe trajectory distortion.
Changing the contact point alone, \texttt{MANOContact} preserves the motion (mean AAE of $1.21^\circ$) yet raises the force RMSE to $5.1\times$ ours: centimeter-level normal errors of the fingertip turn directly into force errors or a loss of contact, and the free-form law absorbs them to match the observed motion, deviating from the actual resistance of the object.
Even though it learns from the measured force, \texttt{ConstantLaw} loses both the motion and the force (mean AAE of $10.89^\circ$, force RMSE $8.4\times$ ours), because a constant law averages out the latch peak, so the microwave door passes its latch without resistance, and cannot supply the return-spring torque that balances gravity, so the oven door falls open.
All four designs are therefore needed to recover accurate joint dynamics.

\subsection{The Learned Law Across Simulators}
\label{sec:exp_sim2sim}

The resistance law is learned in our particle simulator through a compliant finger pad.
To be used in robotics, it should be a standard joint model that does not depend on how our simulator handles contact and integration and that executes directly in a standard robot simulator.
We therefore plan unseen actions on our twin and execute each plan both in our simulator and in the standard robot simulator Newton~\cite{newton2025}, comparing the resulting motion and contact force.

\noindent\textbf{Setup.}
We use the twins reconstructed in \cref{sec:exp_recon} from the human closing takes of the four objects (\textit{microwave close~1}, \textit{laptop close}, \textit{oven close}, and \textit{drawer close}).
On each twin, the planner optimizes with the cross-entropy method (CEM) a 4\,s push force profile at the contact point (followed by 1\,s without force) that closes the part.
We plan with three force penalty weights $w_f \in \{0.1, 1, 10\}$ per object, and the penalty discourages the plan from applying more force than the part requires.
A simulated Franka arm under a Cartesian impedance controller executes each plan on the twin (Ours) and on a standard joint in Newton with our reconstructed kinematics, joint limits, and learned resistance law.
At each simulation step of Newton, we set the joint torque to $\tau_{\text{grav}}(q) + g(q)$, the Coulomb friction to $f(q)$, and the viscous damping to $c$.
Each plan is computed once and executed open loop, so any discrepancy comes from how the part and its contact are simulated and no replanning corrects for it (details in \cref{sec:supp_sim2sim}).

\begin{table}[t]
  \caption{\textbf{Cross-simulator comparison of the learned law.} Three plans per object under different force penalties $w_f$. Peak force is the highest contact force in our simulator before the part closes. AAE ($^\circ$, cm for the drawer) and generalized force RMSE (N$\cdot$m, N for the drawer) compare our simulator with Newton, averaged over the three plans. Per-plan results are in the supplementary (\cref{sec:supp_sim2sim}).}
  \vspace{-5pt}
  \centering
  \footnotesize
  \setlength{\tabcolsep}{0pt}
  \renewcommand{\arraystretch}{0.92}
  \begin{tabular*}{\linewidth}{@{\extracolsep{\fill}}l ccc cc}
    \toprule
    & \multicolumn{3}{c}{Peak force (N)} & \multicolumn{2}{c}{Ours vs.\ Newton} \\
    \cmidrule{2-4}\cmidrule{5-6}
    Object & $w_f = 0.1$ & $w_f = 1$ & $w_f = 10$ & AAE$\downarrow$ & RMSE$\downarrow$ \\
    \midrule
    Microwave & 5.94 & 6.26 & 6.23 & 0.06 & 0.04 \\
    Laptop    & 2.77 & 2.91 & 3.02 & 0.31 & 0.01 \\
    Oven      & 3.11 & 3.10 & 3.12 & 0.77 & 0.05 \\
    Drawer    & 15.70 & 10.01 & 6.70 & 0.06 & 0.33 \\
    \bottomrule
  \end{tabular*}
  \label{tab:sim2sim}
  \vspace{-15pt}
\end{table}

\noindent\textbf{Results.}
All 12 plans close the part and keep it closed in both simulators (\cref{tab:sim2sim}).
Newton closely reproduces our rollouts despite differences in contact handling and integration, with a mean AAE below $0.8^\circ$ ($0.06\text{\,cm}$ for the drawer) and a generalized force RMSE of at most $0.05\text{\,N}\cdot\text{m}$ ($0.33\text{\,N}$ for the drawer), and opening the parts with a grasping tool is equally consistent (\cref{sec:supp_sim2sim}).
This confirms that the learned law acts as a simulator-independent joint model, so plans optimized on our twin transfer directly to Newton.

The learned law also supports force-aware planning that adapts to each object's mechanism.
Raising the force penalty $w_f$ from $0.1$ to $10$ halves the peak force on the drawer from $15.70\text{\,N}$ to $6.70\text{\,N}$, as the planner slows the motion to reduce resistance from its velocity-dependent soft-close damper.
The peak forces on the other objects remain nearly constant, since the planner still supplies the threshold force set by their position-dependent mechanisms, such as the $\sim 6\text{\,N}$ needed to pass the microwave latch.

\subsection{Generalization to an Unseen Contact Point}
\label{sec:exp_cosim}

\begin{figure}[t]
  \centering
  \includegraphics[width=\linewidth]{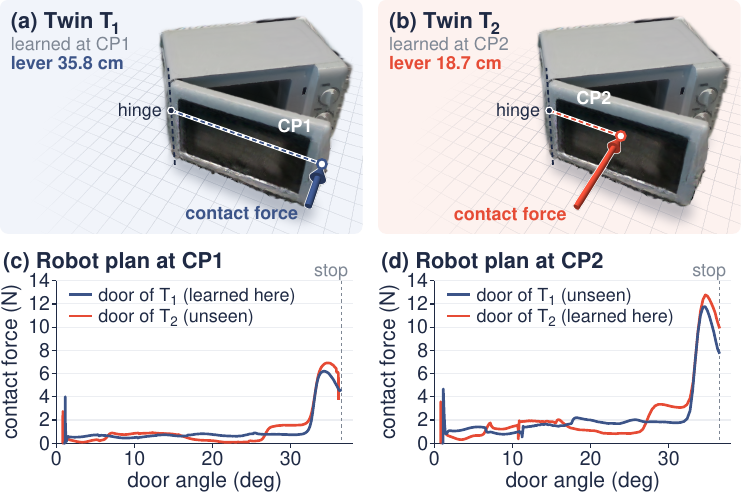}
  \caption{\textbf{Cross-contact validation of the two twins.} (a, b) The interactions at CP1 and CP2 from which $\mathcal{T}_1$ and $\mathcal{T}_2$ are learned. (c, d) Contact force over door angle of both twins under the robot plans computed on $\mathcal{T}_1$ at CP1 and at CP2. Both twins predict about twice the latch force at CP2 as at CP1, also at their unseen contact point.}
  \label{fig:cosim}
  \vspace{-5pt}
\end{figure}

A single interaction manipulates the part at one contact point, yet the full part surface is reconstructed, so a force-grounded twin should predict the response to unseen actions and support planning anywhere on it.
We test this on the microwave door, whose wide face can be manipulated at different lever arms and whose sharp latch peak makes the resulting change in contact force visible.
Passing the latch takes the same torque at any contact point, so the contact force it requires should double at half the lever arm.
If the learned law captures joint dynamics independent of the contact point rather than overfitting to the contact point and force magnitude of its interaction, a twin trained at one contact point should generalize zero-shot to another.

\begin{table}[t]
  \caption{\textbf{Cross-contact comparison of the two twins.} Each twin is trained at one contact point (Train) and tested at both (Test), and zero-shot rows are shaded. The human replay is compared with the measured interaction. In the robot plan, the twin trained at Train plans at Test, and both twins execute it for comparison. AAE in degrees and RMSE in N$\cdot$m.}
  \vspace{-5pt}
  \centering
  \footnotesize
  \setlength{\tabcolsep}{0pt}
  \renewcommand{\arraystretch}{0.92}
  \newcommand{\zsrow}{\smash{\rlap{\textcolor{black!9}{\rule[-\arraystretch\dp\strutbox]{\linewidth}{\dimexpr\arraystretch\ht\strutbox+\arraystretch\dp\strutbox\relax}}}}}
  \begin{tabular*}{\linewidth}{@{\extracolsep{\fill}}l ccc ccc}
    \toprule
    & \multicolumn{3}{c}{Human replay} & \multicolumn{3}{c}{Robot plan} \\
    \cmidrule{2-4}\cmidrule{5-7}
    Train $\rightarrow$ Test & AAE$\downarrow$ & RMSE$\downarrow$ & $r\uparrow$ & AAE$\downarrow$ & RMSE$\downarrow$ & $r\uparrow$ \\
    \midrule
    CP1 $\rightarrow$ CP1 & 0.11 & 0.06 & 1.00 & 0.44 & 0.16 & 0.94 \\
    \zsrow CP2 $\rightarrow$ CP1 & 0.51 & 0.23 & 0.96 & 0.26 & 0.17 & 0.94 \\
    \addlinespace[2pt]
    CP2 $\rightarrow$ CP2 & 0.72 & 0.11 & 0.99 & 0.09 & 0.20 & 0.92 \\
    \zsrow CP1 $\rightarrow$ CP2 & 0.86 & 0.23 & 0.96 & 0.41 & 0.15 & 0.84 \\
    \bottomrule
  \end{tabular*}
  \label{tab:cosim}
  \vspace{-15pt}

\end{table}

\noindent\textbf{Setup.}
From two interactions at the contact points CP1 (lever arm: 35.8\text{\,cm}, 6.7\,N to pass the latch) and CP2 (lever arm: 18.7\text{\,cm}, about half the lever arm and more than twice the force, 15.2\,N), we independently train twins $\mathcal{T}_1$ and $\mathcal{T}_2$ (\cref{fig:cosim}a, b).
Each twin is tested at both CP1 and CP2 (\cref{tab:cosim}) with the AAE, RMSE, and $r$ of \cref{sec:exp_settings}.
At its training contact point, a twin shows how accurately it reconstructs the interaction, and at the other contact point, it is tested zero-shot.
We first replay the human action of each interaction on the twin and compare with the measured motion and force.
We then plan one robot closing action on each twin at each contact point with the CEM planner, and a simulated Franka arm in Isaac Lab~\cite{mittal2025isaaclab} with Newton~\cite{newton2025} executes every plan open loop on both twins via step-by-step co-simulation with our simulator (details in \cref{sec:supp_cosim}).
Since no human measurement exists for the robot plans, the twin trained at each contact point serves there as the reference.

\noindent\textbf{Results.}
As shown in \cref{tab:cosim}, each twin generalizes zero-shot to the unseen contact point almost as well as the twin trained there.
Under the human replay, its predicted force correlates with the measurement at $r = 0.96$, close to $r \geq 0.99$ at the training contact point, and under the robot plans the two twins closely match in both motion and force.

This consistency is most noticeable at the latch (\cref{fig:cosim}c, d).
As expected from the lever arm, both twins predict that passing the latch takes about twice the contact force at CP2 as at CP1 (10.6--12.2\,N \vs 6.0--6.9\,N), also at their unseen contact point.
This shows that a twin learned from one interaction captures joint dynamics independent of the contact point, which transfer zero-shot to force prediction and planning at new contact points.

\subsection{Practical Cost}
\label{sec:exp_cost}

Capturing a twin requires only a single interaction of 5.8 to 10.5\,s, recorded by the calibrated cameras.
On one NVIDIA RTX 5090, the reconstruction takes 71\,min per interaction on average over the four objects, of which preprocessing takes 15\,min, the initial twin 30\,min, and the 1000 steps of the resistance law 26\,min (\cref{sec:supp_cost}).
The finished twin runs at interactive rates: for the action of one frame at 30\,Hz, it returns the motion and the contact force in 42\,ms and the rendered frame in 2 to 5\,ms more, and one plan of \cref{sec:exp_sim2sim} takes 63 to 91\,s on the CPU.

\section{Conclusion}
\label{sec:conclusion}

We presented \method, which reconstructs force-grounded digital twins of articulated objects from a single natural visuo-haptic hand interaction.
By measuring the fingertip force, constraining the contact point to the reconstructed part surface, and learning the resistance law through differentiable rollouts, it reduces sensitivity to observation errors.
Across nine interactions with four real objects, the learned laws capture resistance effects of internal mechanisms, and the twins predict contact forces with $2.9\times$ lower RMSE on average than the strongest baseline.
The twins also demonstrate transfer to an unseen contact point and to a standard robot simulator.
\method advances articulated reconstruction toward twins that reproduce resistance, predict interaction forces, and support force-aware planning beyond the demonstration.

\noindent\textbf{Limitations and future work.}
Although our sensor measures 3D forces, our contact model uses a single pad on a 1-DoF joint, which does not yet cover multi-finger grasps (such as twisting a doorknob) or multi-DoF and coupled joints.
Extending the contact to multiple fingers with a tactile glove and the law to such joints is a natural next step.
In addition, our reconstruction relies on static calibrated RGB-D cameras and learns each law from a single interaction.
Incorporating stronger reconstruction priors for casual videos and combining multiple interactions over a wider range of joint coordinates and velocities would enable in-the-wild capture.

{
    \small
    \bibliographystyle{ieeenat_fullname}
    \bibliography{main}
}

\clearpage
\maketitlesupplementary
\appendix

\noindent\textbf{Supplement Index}\par\smallskip
\noindent\ref{sec:supp_recon}.~Geometric and Kinematic Reconstruction \dotfill \pageref{sec:supp_recon}\\
\ref{sec:supp_learning}.~Resistance-Law Learning in Simulation \dotfill \pageref{sec:supp_learning}\\
\ref{sec:supp_exp_recon}.~Twin Reconstruction from One Interaction \dotfill \pageref{sec:supp_exp_recon}\\
\ref{sec:supp_sim2sim}.~The Learned Law Across Simulators \dotfill \pageref{sec:supp_sim2sim}\\
\ref{sec:supp_cosim}.~Generalization to an Unseen Contact Point \dotfill \pageref{sec:supp_cosim}\\
\ref{sec:supp_cost}.~Practical Cost \dotfill \pageref{sec:supp_cost}\par\medskip

\noindent Our project page (\projectpage) hosts an interactive demo of our four twins and a video.
The demo supports interactive manipulation of each twin through mouse dragging and visualizes the resulting motion and contact forces.

This document also provides further details on the reconstruction and law-learning stages of \method, the extended results for each experiment in the main paper, and the practical computational cost of our pipeline.
Code and data will be released on the project page.
Citation numbers refer to the reference list of the main paper.

\section{Geometric and Kinematic Reconstruction}
\label{sec:supp_recon}

\noindent\textbf{Gaussian initialization.}
We initialize the body Gaussians of $\mathcal{G}$ from the generated object and the part Gaussians from the observed depth.
We generate the object in its open state from all cameras~\cite{li2026mvsam3d} and fit its pose and scale to the object mask and the observed depth.
For the microwave and the drawer, the open part hides surfaces of the body, such as the front frame behind the door and the cabinet face around the drawer, so we also generate the closed state from a single view~\cite{sam3d2025} and fit it in the same way.
The two states expose complementary surfaces, and we merge them at $q = 0$ by the joint $(\mathbf{a}, \mathbf{o})$.
For the part Gaussians, we back-project the depth of the part in each frame $t$, move it from $q^{\text{vis}}_t$ to $q = 0$ along the joint $(\mathbf{a}, \mathbf{o})$, and fuse all frames in 2\,mm voxels. The fused depth leaves holes on glass and on the sides that no camera observes, and we fill them with the generated Gaussians of the part that lie more than 6\,mm from every fused point. These Gaussians only complete the initial geometry and are trained on the images in the same way as those from the depth.

\noindent\textbf{Gaussian training.}
We build on 3DGS~\cite{kerbl20233dgs}, whose L1 and D-SSIM error forms $\mathcal{L}_{\text{color}}$, and densify the Gaussians during 10,000 training iterations.
We train $\mathcal{G}$ together with the joint $(\mathbf{a}, \mathbf{o})$ and the per-frame joint coordinate $q^{\text{vis}}_t$, so that the appearance and the kinematics are consistent.
Each iteration renders one training view with the part posed by the joint at $q^{\text{vis}}_t$, and every third frame is held out.
The cameras observe the object partially, so we regularize the unobserved regions in two ways.
First, in novel views near the cameras, the silhouettes of the generated meshes bound the rendered Gaussians.
Second, after most of the training, we render novel views turned away from the cameras, restore them with GeoQuery~\cite{cao2026geoquery}, and supervise the Gaussians with these restored views under L1 and LPIPS~\cite{zhang2018lpips} errors.

\noindent\textbf{Simulation particles.}
The particles $\mathcal{P}$ take the motion of the object from its 3D tracks and the surface from the Gaussians $\mathcal{G}$.
We track 5{,}000 pixels sampled in the object mask of the first frame of each camera with CoTracker3~\cite{karaev2024cotracker3}, lift them to 3D with depth, and merge the cameras in the world frame.
Every 40 frames, we seed up to 2{,}000 new pixels per camera where the object mask has no visible track, so that the surface revealed by the motion, such as the back of a door, is also tracked.
We drop a track in the frames where it leaves the object mask and remove the points whose motion differs from that of their neighbors within 1\,cm.
We label each point as part or body by whether its track follows the point moved by the joint to $q^{\text{vis}}_t$ or the point held still.
When both fit about equally, the nearer opaque Gaussians decide, and otherwise the segmentation label.
We remove the points far from the opaque Gaussians, which come from failed segmentations.
Where the tracks miss the surface of the part, we add particles at its opaque Gaussians.
The part particles share a nominal mass, and their distribution about the joint gives $I$ and $\tau_{\text{grav}}$.

\section{Resistance-Law Learning in Simulation}
\label{sec:supp_learning}

\subsection{Resistance-Law Initialization}
\label{sec:supp_law}

We initialize $\Theta$ as a standard simulator joint with constant friction $f_0$, damping $c$, and $g(q) = 0$ (\cref{sec:learning}).
Both MLPs take $q$ normalized to the range of motion, and their output layers start at zero, so that $V(q)$ starts flat and $g(q) = 0$.
An output offset $b_0$ with $\operatorname{softplus}(b_0) = f_0$ lets $f(q)$ start at the constant $f_0$.
We estimate $f_0$ from the hand-free motion, the frames where the force sensor reports no contact while the part moves.
We keep runs of at least five frames that travel at least $2^\circ$ (2\,cm for the drawer), which excludes tracking jitter.
On these frames, $\tau_c = 0$ in \cref{eq:eom}, so gravity and inertia give the joint resistance, and $f_0$ is the median of $\operatorname{sgn}(\dot{q})\,(\tau_{\text{grav}}(q) - I\ddot{q})$ over them.
The sensor resolves 0.1\,N, so $f_0$ is at least the torque of this force at the median lever arm of the pad, or 0.1\,N on the drawer.
When the part never moves without the hand, $f_0$ takes this minimum.
The damping starts at a small $c = 0.01$\,N$\cdot$m$\cdot$s, so that friction carries the initial resistance, and a softplus keeps $c \ge 0$ during learning.

\subsection{Fingertip Contact}
\label{sec:supp_contact}

\noindent\textbf{Pad model.}
On a revolute part, the contact point $\mathbf{p}_t$ is the midpoint of the thumb and index fingertips of the fitted MANO hand, since the two are rarely occluded together, and on a prismatic part, it is the pad of the finger that wears the sensor.
We project this point onto the contact face of the reconstructed part along its normal, which keeps its position within the face and discards its depth, the direction in which the fitted hand is least reliable and the part moves.
Only the force component along $\mathbf{J}_t$ does work on the joint, and $\mathbf{J}_t$ is normal to the contact face, so the pad models the normal force alone.
With the compression $\delta_t = \rho_t \sin\big(s_t(\varphi_t - q)\big)$ on a revolute joint and $\delta_t = s_t(\varphi_t - q)$ on a prismatic joint, the normal force is
\begin{equation}
  F_n = \min\Big(\Big[k_n\,\delta_t + c_n\,s_t\,\rho_t\,(\dot{\varphi}_t - \dot{q})\Big]_{+},\, F_{\max}\Big),
  \label{eq:padforce}
\end{equation}
where the upper bound $F_{\max}$ keeps the force and its gradient bounded when a rollout leaves the observed motion.
The pad belongs to the finger, so we use $k_n = 250$\,N/m, $c_n = 0.01\,\text{s} \times k_n = 2.5$\,N\,s/m, and $F_{\max} = 50$\,N for all objects.
A stiff pad would hold the part on the observed motion under any law, so the pad has to be soft: at 250\,N/m, a law that is off by 1\,N shifts the part by 4\,mm, which $\mathcal{L}_{q}$ and $\mathcal{L}_{\text{img}}$ penalize, and a tracking error of 1\,mm changes the force by only 0.25\,N.
The damping term of \cref{eq:padforce} contributes 2--4\% of the measured force, so the pad force is dominated by its elastic term.
The upper bound $F_{\max}$ lies above the largest measured force (15.2\,N) and above every pad force in the rollouts of \cref{tab:main} (at most 22.1\,N), so it never clips the force in our experiments.

The finger position $\varphi_t$ of \cref{eq:pad} is built on $\bar{q}^{\text{vis}}_t$, the per-frame joint coordinate after smoothing, held at its median while the hand exerts force on a part that stays still, and kept inside the joint stops.
While the pad is in contact below $F_{\max}$, we drop the damping term and, on a revolute joint, linearize $\delta_t$ in the small angle $s_t(\varphi_t - q)$, so that $\delta_t = s_t\rho_t(\varphi_t - q)$ on both joints with $\rho_t = 1$ on a prismatic one.
Substituting \cref{eq:pad} then gives the generalized contact force
\begin{equation*}
  \tau_c = s_t\,\rho_t\,F^{\text{J}}_t + k_n\,\rho_t^{2}\,\big(\bar{q}^{\text{vis}}_t - q\big),
\end{equation*}
so the pad applies the measured force plus a spring of stiffness $k_n\rho_t^2$ centered at $\bar{q}^{\text{vis}}_t$.
At $q = \bar{q}^{\text{vis}}_t$, the elastic contribution of the pad thus matches the measured force, and holding $\bar{q}^{\text{vis}}_t$ on a still part keeps tracking jitter out of $\tau_c$.

\noindent\textbf{Measured force.}
The sensor reports $\mathbf{F}_t$ in its own frame without precise pose information.
On a revolute part, the finger contacts the face, so we align the compression axis of the sensor with the face normal, rotated with the joint.
On a prismatic part, the finger acts on a handle, and we place the magnitude $\lVert \mathbf{F}_t \rVert$ along $s_t\,\mathbf{J}_t$, the direction in which the hand moves the part.
We take the measured force of \cref{eq:pad} along the face normal, which coincides with $F^{\text{J}}_t$ for a face through the joint axis.

The contact frames $\mathcal{C}$ are the frames on which the sensed finger drives the part.
A contact span is a run of frames with a normal reading above the sensor resolution of 0.1\,N.
A span enters $\mathcal{C}$ as a whole when its measured force does positive work on the joint, which also keeps the frames in which the hand pushes the part against a stop.
Outside $\mathcal{C}$ the finger exerts no force.

\subsection{Differentiable Simulation}
\label{sec:supp_diffsim}

\noindent\textbf{Joint stops.}
The simulation confines $q$ to the range of motion $[q_{\min}, q_{\max}]$ and removes the velocity into either end.
An end is a stop when the interaction shows that the static body holds the part there, that is, the part rests at its extreme coordinate without the hand, and at this coordinate the hand pushes on the part while it stays still or the part arrives fast and halts within a few frames.
The stop sits at the median $q^{\text{vis}}_t$ of these frames, and an end without a stop lies a margin beyond the observed $q^{\text{vis}}_t$.
A stop can hold the part against a nonzero net conservative torque $N(q)$, which $\mathcal{L}_{\text{rest}}$ accounts for below.

\noindent\textbf{Rollout length and loss scales.}
A short window keeps the rollout close to the observed motion, and $K = 32$ frames ($\sim$1\,s) still let a torque error integrate into a visible position error, as in a free motion or a spring-back after release.
Every fifth frame is held out, and the loss on these frames selects the final model.

We scale the motion and torque losses of \cref{eq:total} to one unit, the error of a $1^\circ$ offset of the part, or of 1\,cm on a prismatic joint.
$\mathcal{L}_{q}$ is measured in squared degrees, and we estimate $\Delta_{\text{img}}$ before learning from renderings at $q^{\text{vis}}_t$ and $q^{\text{vis}}_t \pm 1^\circ$, so an image misfit of $1^\circ$ weighs as much as a $1^\circ$ error in $\mathcal{L}_{q}$.
$\mathcal{L}_{\text{rest}}$ and $\mathcal{L}_{V}$ divide the torque by the torque scale $\tau_0$, so a torque of $\tau_0$ weighs as much as a $1^\circ$ error.
We set $\tau_0$ to the initial friction $f_0$, which is at least the torque of the sensor resolution (\cref{sec:supp_law}) and thus stays away from zero.

\noindent\textbf{Rest equilibria.}
The simulator smooths $\operatorname{sgn}(\dot{q})$ in \cref{eq:eom} with a $\tanh$, so the friction resists a slow motion like a stiff damper and cannot hold the part against a nonzero net conservative torque $N(q) = g(q) + \tau_{\text{grav}}(q)$.
A simulated part with $N \neq 0$ therefore drifts slowly in the direction of $N$ and rests only where $N = 0$.
The rollout losses barely see this slow drift within $K$ frames.
$\mathcal{L}_{\text{rest}}$ therefore penalizes $N$ itself on the rest frames $\mathcal{R}$.
A stop holds the part against a torque that points into it, so on a stop at $q_{\max}$ ($\sigma = 1$) or $q_{\min}$ ($\sigma = -1$) only a torque away from the stop is an error,
\begin{equation}
  \mathcal{L}_{\text{rest}} = \frac{1}{\tau_0^{2}}\Big\langle
  \begin{cases}
    N(q^{\text{vis}}_{t})^{2} & \text{interior},\\
    \big[\, -\sigma\,N(q^{\text{vis}}_{t}) \,\big]_{+}^{2} & \text{on a stop},
  \end{cases}
  \Big\rangle_{t \in \mathcal{R}}\,.
  \label{eq:rest}
\end{equation}
$\mathcal{L}_{\text{rest}}$ constrains $g(q)$ only at the coordinates where the part rests, and the prior $\mathcal{L}_{V}$ covers the others by averaging $(g(q)/\tau_0)^2$ uniformly over $[q_{\min}, q_{\max}]$.

\noindent\textbf{Training.}
We train $\Theta$ with Adam for 1000 steps at a learning rate of $3 \times 10^{-3}$, with 20 substeps per frame, $\lambda_{q} = 1$, $\lambda_{F} = 10$, $\lambda_{V} = 0.02$, and $\epsilon = 0.5$\,N.
The door of the oven close take carries a spring torque far above its torque scale $\tau_0$, so we train this take for 4000 steps with $\lambda_{V} = 0$ in the last 1000, and \texttt{VisionForce} follows the same schedule.

We trained the Gaussians $\mathcal{G}$ with the part posed at the per-frame joint coordinate $q^{\text{vis}}_t$, and the twin renders the part at the simulated coordinate $\hat{q}_t$, which differs slightly.
Between law updates, we therefore refine the part Gaussians with $\mathcal{L}_{\text{img}}$ at $\hat{q}_t$, keeping the joint $(\mathbf{a}, \mathbf{o})$ fixed and the Gaussian centers within a few millimeters of the reconstruction.

\section{Twin Reconstruction from One Interaction}
\label{sec:supp_exp_recon}

\subsection{Baselines and Ablations}
\label{sec:supp_baselines}

Four baselines adapt existing lines of work to our setting, and two ablations each change one design of our own method (\cref{sec:exp_recon}).
All variants keep the joint $(\mathbf{a}, \mathbf{o})$, the particles $\mathcal{P}$, and the Gaussians $\mathcal{G}$ of our method, and all are scored against the sensor.
Below, we detail each adaptation.

\noindent\textbf{(A) Vision-based force (\texttt{VisionForce}).}
EgoTac~\cite{zhang2026egotac} predicts the normal force on every MANO vertex from RGB video of the hand, so this baseline tests whether a visual prior can replace the direct force measurement.
We run its released video model, which takes four frames as context (checkpoint \texttt{egotac\_base\_h4}), zero-shot on a hand-centered crop of the camera that sees the hand best and sum the prediction over the sensed fingertip.
This magnitude replaces the sensor magnitude on the contact frames $\mathcal{C}$, along the force direction of the sensor, and our method, including the stop detection, then runs unchanged.

\noindent\textbf{(B) Generated asset dynamics (\texttt{AssetGen}).}
Artisan~\cite{cai2026artisan} generates simulation assets with agents that assign the part mass and constant joint friction and damping from the appearance of the object, so this baseline tests whether a foundation-model prior can replace the direct force measurement.
We generate one asset per object with the configuration of the original paper, from a frame of a take and a text that describes the appearance, the moving part, the joint type, and the range of motion (\cref{tab:supp_assetgen}).
The generated asset differs from the observed object in shape and texture, so \texttt{AssetGen} keeps our reconstruction and takes only the dynamics of the asset.
These dynamics, namely its Coulomb friction, viscous damping, inertia, and the gravity torque of its mass about our joint, replace the rigid-body terms and the resistance law of \cref{eq:eom}, and the finger action $\mathbf{u}_t$, the fingertip contact, and the joint stops stay those of our method.
The assigned parameters are constant over the joint coordinate and include no spring, and the drawer is assigned zero friction and damping.

\begin{table*}[t]
  \caption{\textbf{Inputs and outputs of \texttt{AssetGen}.} For each object, the reference image and the text that the generator receives, the generated asset and our reconstruction rendered over the reference image, and the mass of the moving part and the constant joint parameters that the generated asset assigns. Friction is in N$\cdot$m and damping in N$\cdot$m$\cdot$s/rad (N and N$\cdot$s/m for the drawer), and the joint stiffness is zero on all four assets. The range of motion is used only when the complete asset is rolled out (\cref{sec:supp_results}).}
  \vspace{-5pt}
  \centering
  \footnotesize
  \setlength{\tabcolsep}{4pt}
  \begin{tabular*}{\linewidth}{@{\extracolsep{\fill}} l c c c c c c c @{}}
    \toprule
    Object & Reference Image & Generated Asset & \method (Ours) & Mass (kg) & Friction & Damping & Range \\
    \midrule
    \addlinespace[2pt]
    Microwave & \raisebox{-0.5\height}{\includegraphics[width=0.18\linewidth]{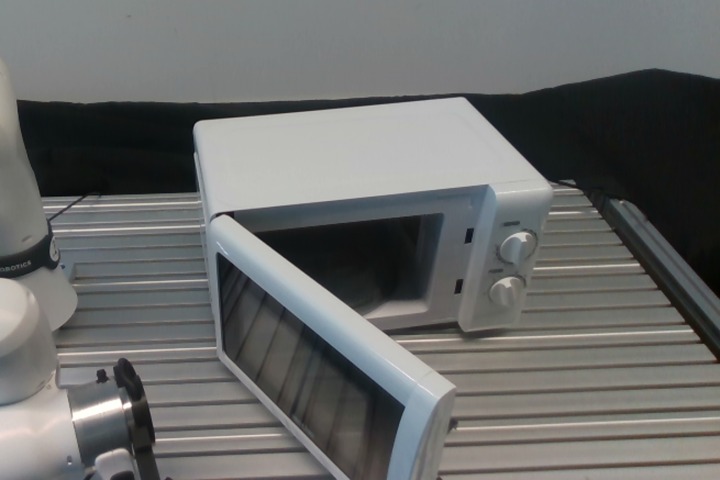}} & \raisebox{-0.5\height}{\includegraphics[width=0.18\linewidth]{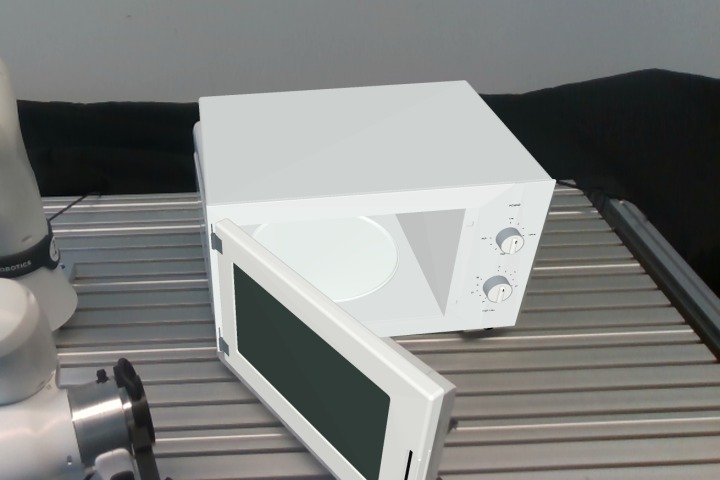}} & \raisebox{-0.5\height}{\includegraphics[width=0.18\linewidth]{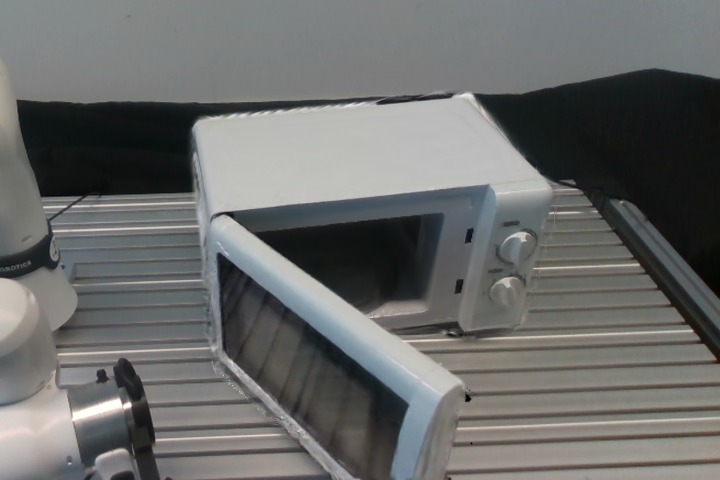}} & 1.15 & 0.075 & 0.043 & 0--100$^\circ$ \\[3pt]
    \multicolumn{8}{@{}p{\linewidth}@{}}{\scriptsize\textit{Text input:} A compact white countertop microwave oven with a boxy, slightly rounded glossy white body. Focus only on the microwave; ignore the robot arm, hand, table and background in the reference image. The front is split into a large door on the left (about three quarters of the width) and a narrow vertical control panel on the right. The door has a white frame with a dark see-through mesh window in the centre and a recessed grip on its free right-hand edge. The control panel carries two silver-and-white round rotary knobs stacked vertically (an upper power-level knob and a lower timer knob), with no digital display or keypad. Inside, the cavity has white walls and a round glass turntable plate on the floor. The door swings outward on a vertical hinge along its left-hand edge, opening from closed to roughly 100 degrees to reveal the cavity. Each of the two knobs rotates about its own axis, perpendicular to the front panel, and the glass turntable plate rotates freely about the vertical axis at the centre of the cavity.} \\
    \midrule
    \addlinespace[2pt]
    Laptop & \raisebox{-0.5\height}{\includegraphics[width=0.18\linewidth]{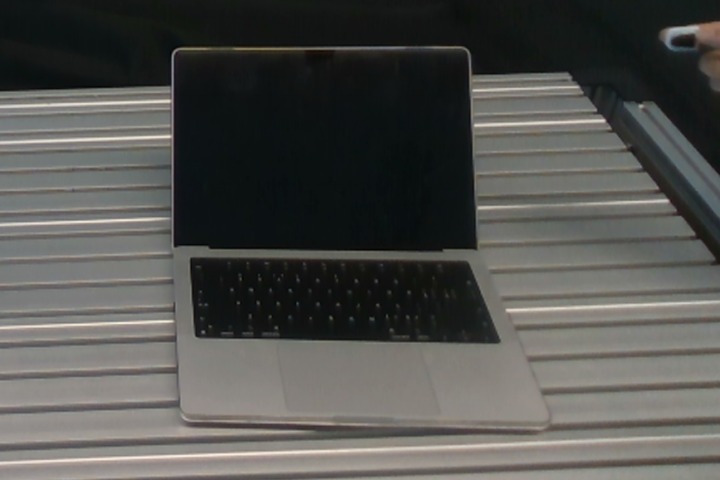}} & \raisebox{-0.5\height}{\includegraphics[width=0.18\linewidth]{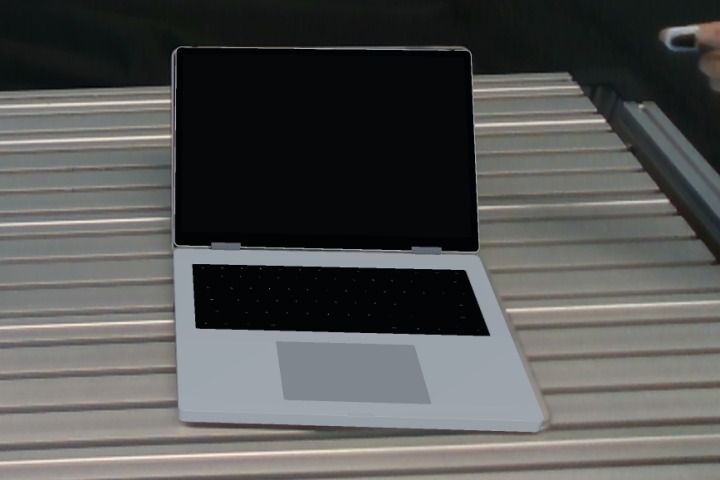}} & \raisebox{-0.5\height}{\includegraphics[width=0.18\linewidth]{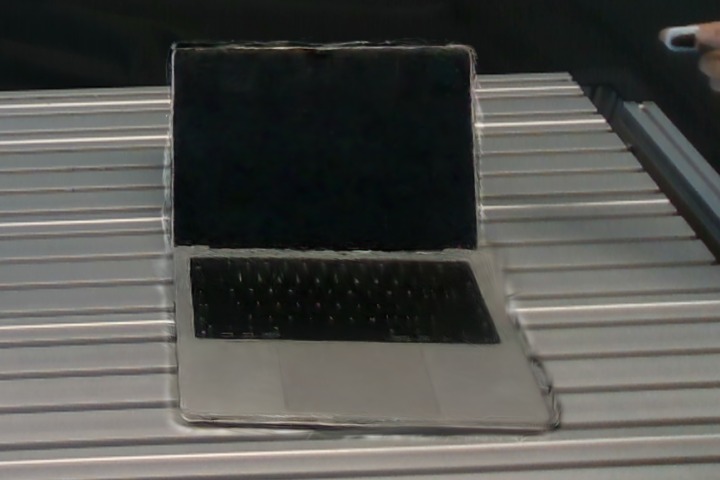}} & 0.32 & 0.433 & 0.005 & 0--130$^\circ$ \\[3pt]
    \multicolumn{8}{@{}p{\linewidth}@{}}{\scriptsize\textit{Text input:} A thin silver aluminium laptop with a dark keyboard and a large trackpad. Focus only on the laptop; ignore the robot arm, hand, table and background in the reference image. The laptop consists of a flat base that rests on the table and a thin lid that carries a black display with a narrow bezel. The base carries a full black keyboard in its rear part and a wide trackpad centred in front of the keyboard. The lid rotates on a horizontal hinge along the rear edge of the base, from closed (lying flat on the keyboard) to roughly 130 degrees open.} \\
    \midrule
    \addlinespace[2pt]
    Oven & \raisebox{-0.5\height}{\includegraphics[width=0.18\linewidth]{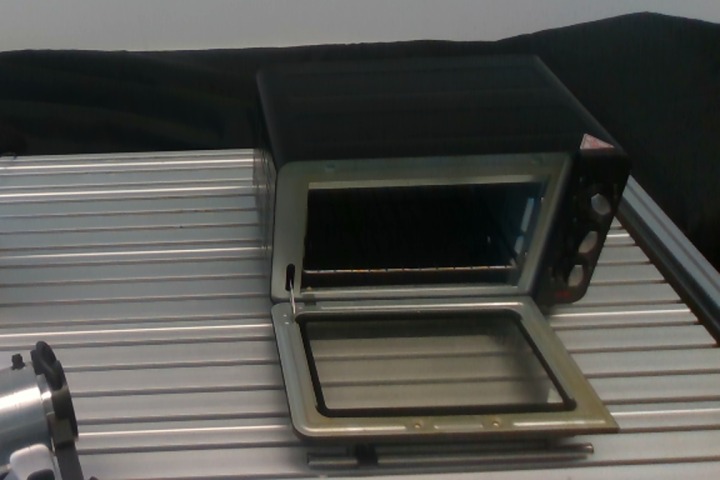}} & \raisebox{-0.5\height}{\includegraphics[width=0.18\linewidth]{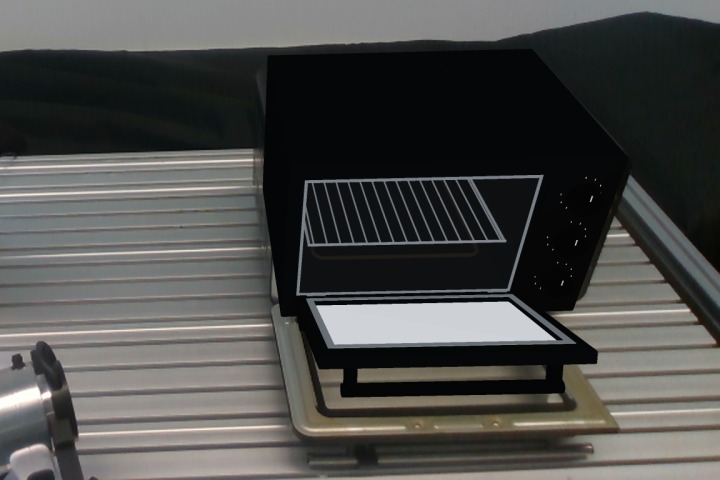}} & \raisebox{-0.5\height}{\includegraphics[width=0.18\linewidth]{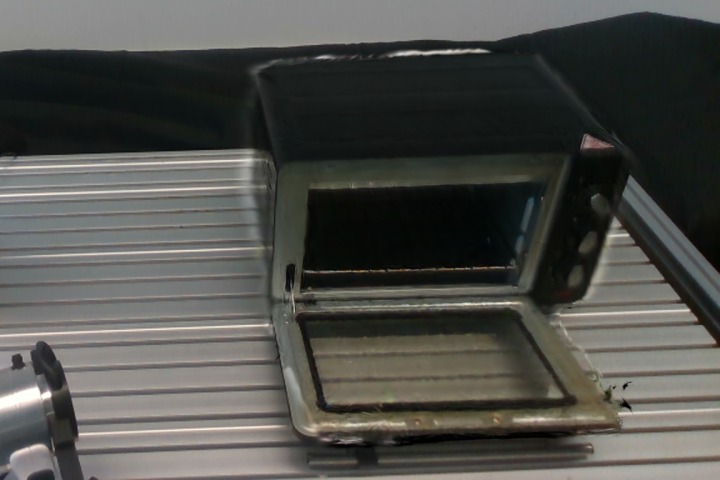}} & 0.92 & 0.094 & 0.044 & 0--90$^\circ$ \\[3pt]
    \multicolumn{8}{@{}p{\linewidth}@{}}{\scriptsize\textit{Text input:} A compact black countertop toaster oven with a boxy body, a matte black housing and a vented top. Focus only on the oven; ignore the robot arm, hand, table and background in the reference image. The front is split into a large door on the left (about three quarters of the width) and a narrow vertical control panel on the right. The door has a black frame with a large clear glass window and a horizontal handle bar along its free top edge. The control panel carries three round black rotary knobs stacked vertically. Inside, the cavity has dark walls, heating elements along the top and bottom, and a wire rack. The door swings downward on a horizontal hinge along its bottom edge, opening from closed (vertical) to roughly 90 degrees (horizontal) to reveal the cavity. Each of the three knobs rotates about its own axis, perpendicular to the front panel.} \\
    \midrule
    \addlinespace[2pt]
    Drawer & \raisebox{-0.5\height}{\includegraphics[width=0.18\linewidth]{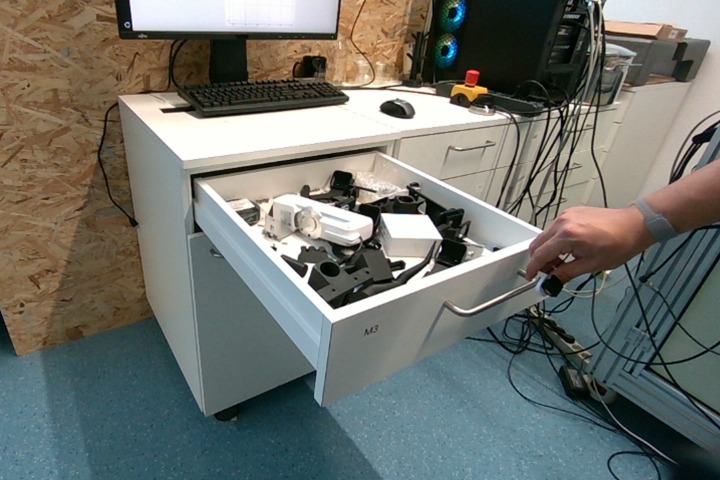}} & \raisebox{-0.5\height}{\includegraphics[width=0.18\linewidth]{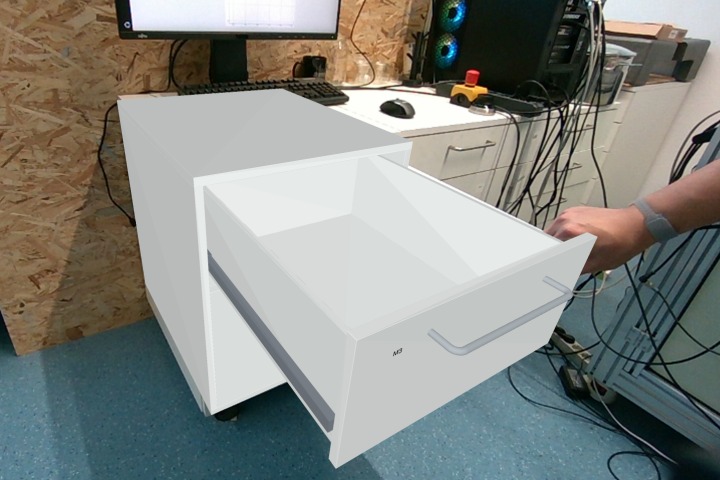}} & \raisebox{-0.5\height}{\includegraphics[width=0.18\linewidth]{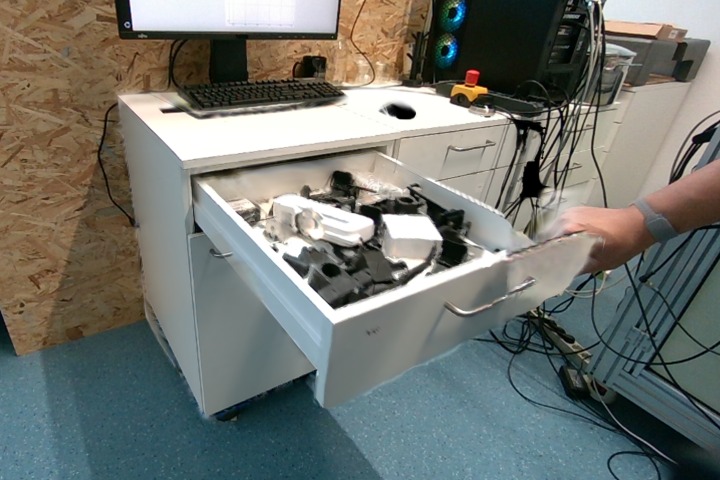}} & 4.00 & 0 & 0 & 0--40\,cm \\[3pt]
    \multicolumn{8}{@{}p{\linewidth}@{}}{\scriptsize\textit{Text input:} A white mobile office pedestal cabinet on small castors with a flat white top. Focus only on the cabinet and its open drawer; ignore the hand, the contents of the drawer, the desk, cables and background in the reference image. The front carries a deep drawer with a white front panel, a horizontal stainless-steel bar handle near its top edge, and a small printed label. The drawer box has white side walls and a flat bottom. The drawer slides straight out of the cabinet on a prismatic joint along the depth direction, from closed (flush with the cabinet front) to roughly 40 centimetres open.} \\
    \bottomrule
  \end{tabular*}
  \label{tab:supp_assetgen}
\end{table*}

\noindent\textbf{(C) System identification (\texttt{SysID}).}
Wrench-based systems identify joint dynamics by regressing the measured torque on the observed motion~\cite{engelbracht2026forcetwin}, and we apply this identification to our data to test whether frame-by-frame regression can replace learning in the simulation loop.
On the contact frames $\mathcal{C}$, we differentiate $q^{\text{vis}}_t$ with a Savitzky-Golay filter of 9 frames and fit the measured generalized force $\mathbf{F}_t^{\top}\mathbf{J}_t$ at $q = q^{\text{vis}}_t$,
\begin{equation}
  \mathbf{F}_t^{\top}\mathbf{J}_t = \tilde{I}\,\ddot{q} + \mu\,\operatorname{sgn}(\dot{q}) + b\,\dot{q} + h(q) + d(q, \lvert\dot{q}\rvert)\,\dot{q} .
  \label{eq:sysid}
\end{equation}
Non-negative least squares gives the inertia $\tilde{I}$, the Coulomb friction $\mu$, and the viscous damping $b$, and two MLPs, a torque $h$ and a damping $d \ge 0$, then fit the residual.
We integrate the identified law under the same finger action $\mathbf{u}_t$, fingertip contact, and joint stops as our twin.
Wider windows of 15, 21, and 31 frames lower the per-frame residual of \cref{eq:sysid} on six to seven of the nine takes, the AAE of the rollout on two to three, and its force RMSE on one to two.
The mean force RMSE rises from $2.9\times$ ours with the 9-frame window (\cref{tab:main}) to $6.2\times$, $7.0\times$, and $8.1\times$.

\noindent\textbf{(D) Video twin with force supervision (\texttt{VideoTwin}).}
PhysTwin~\cite{jiang2025phystwin} drags the object with control springs attached to the tracked hand and learns the physical parameters from video, so this baseline tests whether coupling to the tracked hand can replace our surface-constrained contact.
We keep its hand coupling, with the springs wired at the first frame, and its two training stages of CMA-ES and gradient descent.
We restrict the part to our joint with a constant damping and friction, bound its motion a margin beyond the observed $q^{\text{vis}}_t$, and add to both stages a loss between the resultant spring force on the part and the measured force.

\noindent\textbf{(E) MANO contact point (\texttt{MANOContact}).}
This ablation isolates how the tracking error of the fingertip acts on a rigid part.
It keeps the depth of the fitted MANO fingertip along the face normal, which our method discards (\cref{sec:supp_contact}), so the joint coordinate of this fingertip replaces \cref{eq:pad} as the finger position $\varphi_t$ on every frame.
The contact frames $\mathcal{C}$, the pad model, the resistance law, and the training stay those of our method.

\noindent\textbf{(F) Constant joint law (\texttt{ConstantLaw}).}
This ablation tests whether constant joint parameters can replace the free-form resistance law.
It replaces $f(q)$ and $g(q)$ in \cref{eq:eom} with a constant friction $f$ and a linear spring $g(q) = -k\,q$ about the pose of the first frame, with $k \ge 0$, and learns $f$, $k$, and $c$ with our losses and training.

\subsection{Metrics}
\label{sec:supp_metrics}
We detail how the motion and appearance metrics of \cref{sec:exp_settings} are computed.
AAE is the mean absolute difference between the simulated coordinate $\hat{q}_t$ and $q^{\text{vis}}_t$ over all frames of a take, in degrees, or in cm on the drawer.
CD-m is single-directional, from the observation to the prediction.
For every tracked 3D point of the moving part that is visible in a frame, we take the L1 distance to the nearest surface particle of the part posed at $\hat{q}_t$ and average over the points and the frames, in mm.
PSNR is computed at the main camera over the full image and averaged over the frames, with the Gaussians rendered at $\hat{q}_t$ on a black background, the photo set to black outside the object mask, and the hand removed from both images.
We complete the object mask with the silhouette of the Gaussians rendered at $q^{\text{vis}}_t$, which is the same for every method, because the segmentation misses parts of the object surface.

\subsection{Results}
\label{sec:supp_results}

\begin{table*}[t]
  \caption{\textbf{Motion and appearance on twin reconstruction.} AAE, CD-m, and PSNR of the open-loop rollout of each method and of the ablations (E) and (F) on the nine takes of \cref{tab:main}. AAE in degrees (cm for the drawer), CD-m in mm, and PSNR in dB. The best value of each take is in bold.}
  \label{tab:supp_motion}
  \vspace{-5pt}
  \centering
  \footnotesize
  \setlength{\tabcolsep}{4pt}
  \renewcommand{\arraystretch}{0.92}
  \begin{tabular}{ll ccc cc cc cc}
    \toprule
    & & \multicolumn{3}{c}{Microwave} & \multicolumn{2}{c}{Laptop} & \multicolumn{2}{c}{Oven} & \multicolumn{2}{c}{Drawer} \\
    \cmidrule(lr){3-5}\cmidrule(lr){6-7}\cmidrule(lr){8-9}\cmidrule(lr){10-11}
    Metric & Method & close 1 & close 2 & open & close & open & open & close & close & open \\
    \midrule
    AAE$\downarrow$ & (A) \texttt{VisionForce} & 0.19 & 0.23 & \textbf{0.17} & 0.42 & \textbf{0.37} & \textbf{0.21} & 0.52 & 0.19 & 0.69 \\
     & (B) \texttt{AssetGen} & 0.61 & 5.10 & 9.87 & 6.14 & 1.68 & 31.48 & 27.17 & 2.69 & 18.73 \\
     & (C) \texttt{SysID} & 6.54 & 2.35 & 3.98 & 5.37 & 1.29 & 0.66 & 10.49 & 0.27 & 1.95 \\
     & (D) \texttt{VideoTwin} & 15.62 & 18.63 & 20.61 & 63.02 & 22.37 & 2.78 & 10.32 & 10.69 & 2.32 \\
     & (E) \texttt{MANOContact} & 1.42 & 0.63 & 0.80 & 1.47 & 2.35 & 0.63 & 1.16 & 0.93 & 2.72 \\
     & (F) \texttt{ConstantLaw} & 0.61 & 4.75 & 9.39 & 1.80 & 1.37 & 31.06 & 27.26 & 1.64 & 15.92 \\
     & \method (Ours) & \textbf{0.16} & \textbf{0.14} & 0.72 & \textbf{0.27} & 0.40 & 0.28 & \textbf{0.13} & \textbf{0.14} & \textbf{0.39} \\
    \midrule
    CD-m$\downarrow$ & (A) \texttt{VisionForce} & 4.47 & \textbf{4.69} & 4.14 & 7.47 & 8.38 & \textbf{4.94} & 5.20 & 4.74 & 11.10 \\
     & (B) \texttt{AssetGen} & 4.51 & 17.46 & 31.23 & 10.74 & 7.88 & 127.69 & 34.85 & 28.01 & 150.50 \\
     & (C) \texttt{SysID} & 14.52 & 9.31 & 14.93 & 9.95 & \textbf{7.59} & 5.23 & 16.78 & 6.61 & 23.49 \\
     & (D) \texttt{VideoTwin} & 38.04 & 44.68 & 75.82 & 64.59 & 27.43 & 8.09 & 17.37 & 65.88 & 27.21 \\
     & (E) \texttt{MANOContact} & 6.51 & 4.78 & 4.88 & \textbf{6.67} & 10.37 & 5.34 & 5.40 & 13.20 & 30.48 \\
     & (F) \texttt{ConstantLaw} & 4.55 & 16.06 & 29.49 & 7.56 & 8.14 & 126.38 & 34.88 & 19.81 & 129.66 \\
     & \method (Ours) & \textbf{4.45} & 4.77 & \textbf{4.04} & 6.76 & 8.27 & 5.03 & \textbf{4.86} & \textbf{4.56} & \textbf{9.34} \\
    \midrule
    PSNR$\uparrow$ & (A) \texttt{VisionForce} & 31.79 & 27.71 & 33.00 & 33.28 & \textbf{33.30} & \textbf{27.65} & 26.01 & 23.77 & \textbf{23.26} \\
     & (B) \texttt{AssetGen} & 30.81 & 24.89 & 28.33 & 31.82 & 32.56 & 21.40 & 22.60 & 22.32 & 19.79 \\
     & (C) \texttt{SysID} & 27.40 & 25.66 & 30.36 & 31.70 & 32.71 & 27.64 & 24.19 & \textbf{23.87} & 22.87 \\
     & (D) \texttt{VideoTwin} & 23.29 & 22.39 & 26.48 & 28.38 & 29.22 & 26.98 & 22.58 & 20.72 & 22.51 \\
     & (E) \texttt{MANOContact} & 29.83 & \textbf{27.81} & 32.26 & 33.01 & 32.70 & 27.48 & 25.87 & 23.71 & 22.25 \\
     & (F) \texttt{ConstantLaw} & 31.08 & 25.06 & 28.79 & 32.83 & 32.74 & 21.47 & 22.42 & 23.02 & 20.24 \\
     & \method (Ours) & \textbf{31.83} & 27.70 & \textbf{33.15} & \textbf{33.32} & 33.27 & 27.62 & \textbf{26.08} & 23.86 & 23.24 \\
    \bottomrule
  \end{tabular}
\end{table*}

\begin{figure*}[p]
  \begin{minipage}[c][0.995\textheight][s]{\linewidth}
    \includegraphics[width=\linewidth]{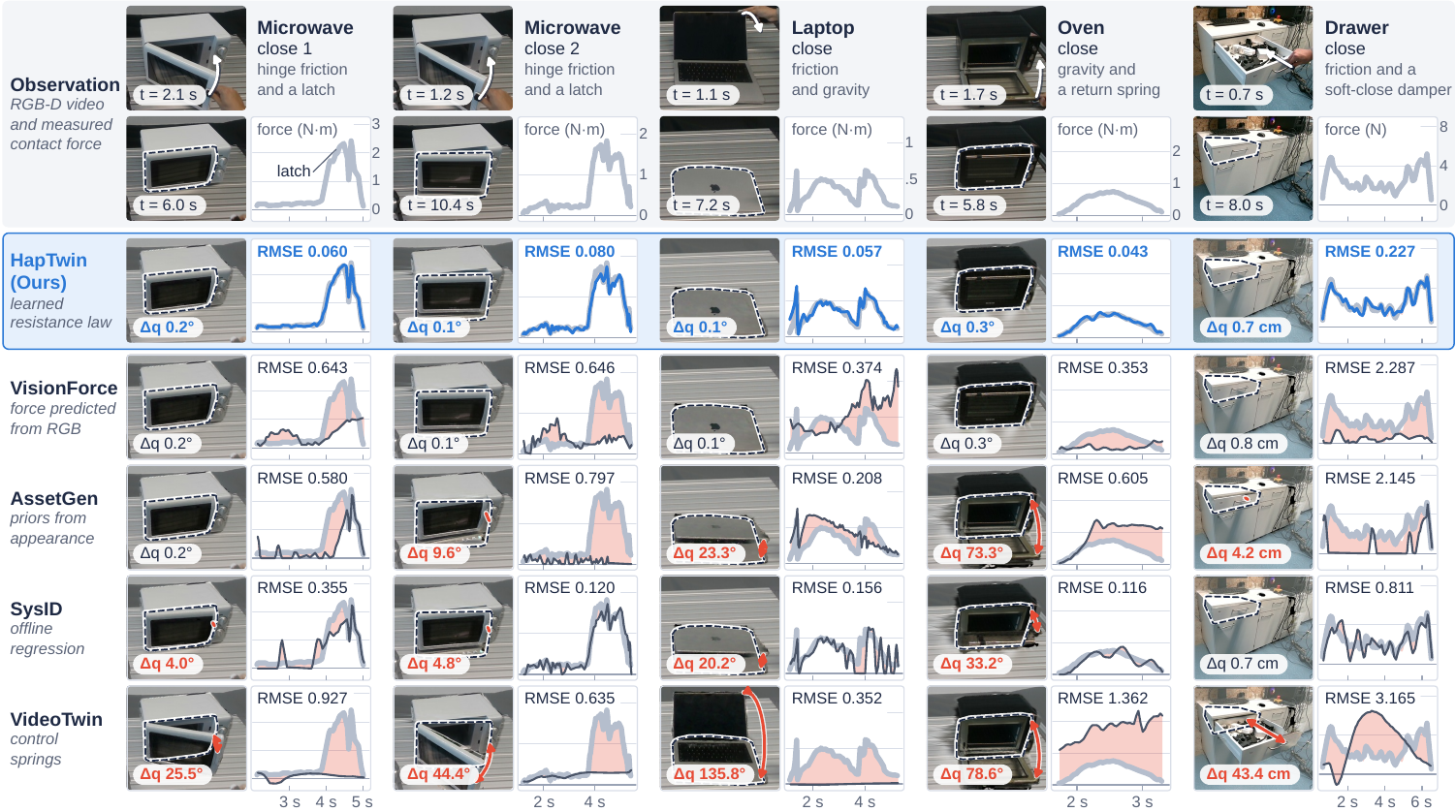}\par
    \vspace{6pt plus 1fill}
    \includegraphics[width=\linewidth]{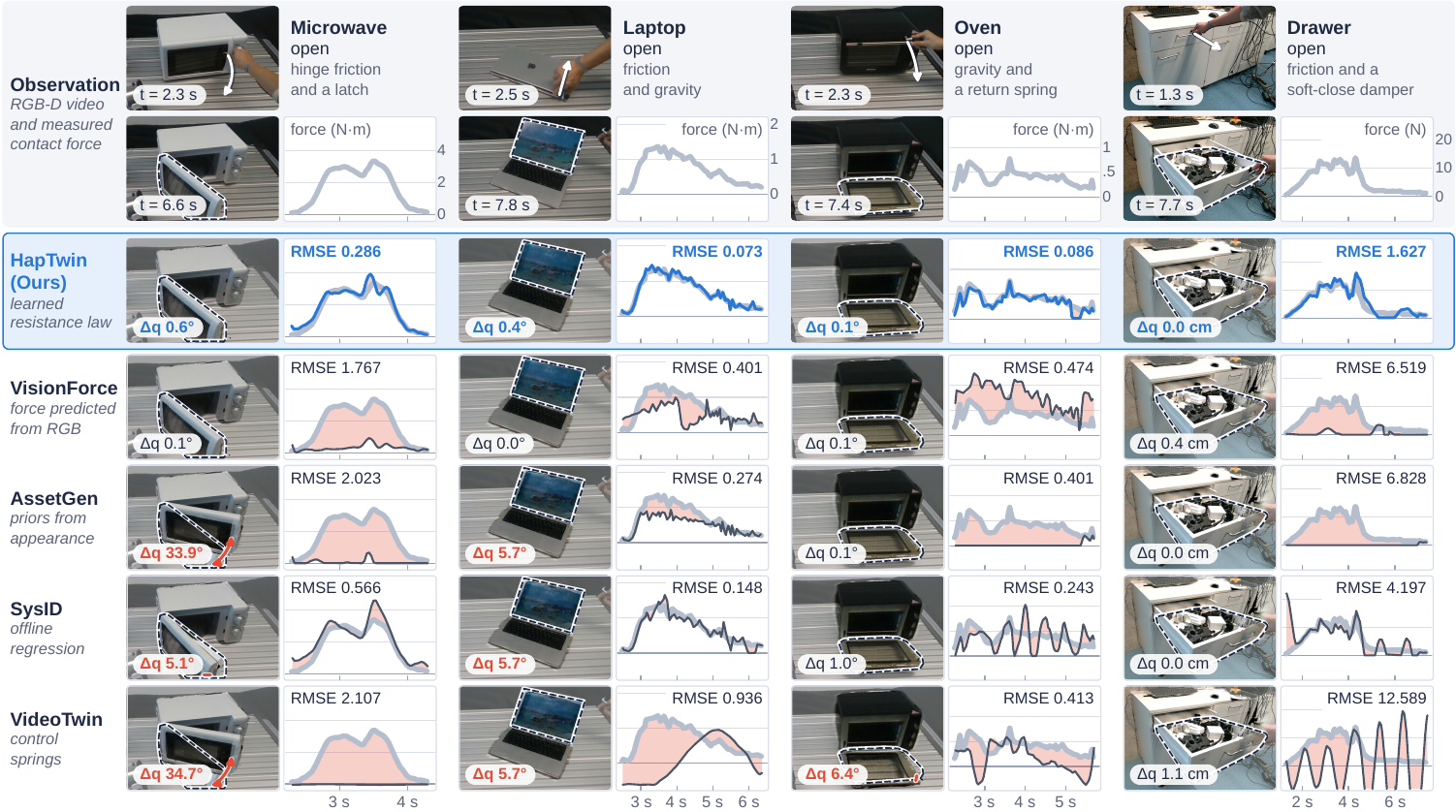}\par
    \vspace{4pt plus 1fill}
    \includegraphics[width=\linewidth]{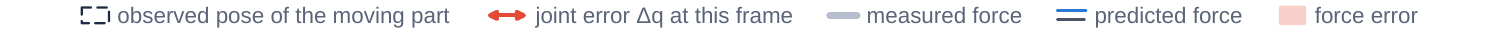}\par
    \caption{\textbf{Qualitative results on all nine takes.} As in \cref{fig:replay}, we visualize the open-loop rollout at the end of the take and the generalized contact force of each method, here on the five closing takes (top) and the four opening takes (bottom). The RMSE values are those of \cref{tab:main}. On \textit{drawer open}, the force predicted by \texttt{VideoTwin} oscillates between $-23$ and $20$\,N, and its curve is cut at the lower edge of the tile.}
    \label{fig:supp_replay}
  \end{minipage}
\end{figure*}

\Cref{tab:supp_motion} gives the per-take values behind \cref{fig:motion}, and \cref{fig:supp_replay} extends \cref{fig:replay} to all nine takes.

On every take, \method and \texttt{VisionForce} are on par in motion and appearance, with an AAE below $1^\circ$ (1\,cm for the drawer) and differences of at most 1.8\,mm in CD-m and 0.2\,dB in PSNR (\cref{tab:supp_motion}), while the force RMSE of \texttt{VisionForce} is $4.0\times$ to $10.7\times$ ours (\cref{tab:main}).
On the three microwave takes, its predicted force peaks at 0.87--1.08\,N$\cdot$m against measured peaks of 1.81--3.32\,N$\cdot$m, so a twin can replay the video and still miss the resistance of the latch.

With one twin per take, the learned laws reproduce the resistance of all four mechanisms, the latch, the friction hinge, the return spring, and the soft-close damper: the predicted force follows the measured force with $r = 0.93$--$1.00$ on the closing takes and $r = 0.89$--$0.99$ on the opening takes, and the part ends within $0.6^\circ$ (0.7\,cm for the drawer) of the observed pose (\cref{fig:supp_replay}).
In \textit{microwave close~2}, the door does not close and swings back open after the hand releases it, and our twin follows this free motion with an AAE of $0.14^\circ$, so the learned law also reproduces the motion of the released part.

\texttt{SysID} comes closest to our force, yet its rollout drifts, with an AAE above $1^\circ$ (1\,cm for the drawer) on seven of the nine takes.
On \textit{oven close}, its force correlates with the measured force at $r = 0.88$ while the door ends $33^\circ$ from the observed closed pose (\cref{fig:supp_replay}), so a law that fits the measured force frame by frame can still miss the motion once it is integrated.

\begin{table}[t]
  \caption{\textbf{Measured force missed by the open-loop rollout.} For each take, we sum the measured generalized force over the contact frames $\mathcal{C}$ and report the share that falls on frames where the predicted force is zero or opposes the hand, averaged over the nine takes of \cref{tab:main}. With a pad contact, these are the frames on which the pad is off the part.}
  \vspace{-5pt}
  \centering
  \footnotesize
  \setlength{\tabcolsep}{8pt}
  \renewcommand{\arraystretch}{0.92}
  \begin{tabular}{l c}
    \toprule
    Method & Missed force (\%)$\downarrow$ \\
    \midrule
    (A) \texttt{VisionForce} & 7.2 \\
    (B) \texttt{AssetGen} & 41.0 \\
    (C) \texttt{SysID} & 4.6 \\
    (D) \texttt{VideoTwin} & 34.0 \\
    \midrule
    (E) \texttt{MANOContact} & 9.5 \\
    (F) \texttt{ConstantLaw} & 33.6 \\
    \midrule
    \method & \textbf{0.7} \\
    \bottomrule
  \end{tabular}
  \label{tab:supp_contact_loss}
\end{table}

With the contact taken from the tracked hand, \texttt{VideoTwin} ends furthest from the observed pose on every closing take, by up to $136^\circ$ on the laptop lid and 43\,cm on the drawer (\cref{fig:supp_replay}), and its springs oppose the hand on frames that carry 34.0\% of the measured force (\cref{tab:supp_contact_loss}).
\texttt{MANOContact} keeps the AAE within $2.4^\circ$ (2.7\,cm for the drawer) on every take, yet its pad is off the part on frames that carry 9.5\% of the measured force, against 0.7\% for our twin, so the error of the tracked fingertip reaches the force as a loss of contact even where the motion is matched.

\texttt{AssetGen} and \texttt{ConstantLaw} both use constant joint parameters, assigned from appearance in the first and learned from the measured force in the second, and their force RMSEs in \cref{tab:main} agree within 5\% on six of the nine takes.
Their predicted peaks on \textit{microwave open} are 0.67 and 0.83\,N$\cdot$m against the measured 3.32\,N$\cdot$m, and on the oven both reach an AAE of $31^\circ$ on \textit{oven open} and leave the door $73^\circ$ from the observed pose at the end of \textit{oven close}.

\texttt{AssetGen} places the generated dynamics on our joint and geometry, and we also roll out the complete generated asset, with its own geometry, joint, and range of motion, registered to each take and driven by the same finger action and fingertip contact.
Its force RMSE is $7.7\times$ ours on average over the nine takes, against $7.4\times$ for \texttt{AssetGen} in \cref{tab:main}, so the force error of this baseline persists with its own geometry.
The generated oven has no return spring, so its door falls open under its own weight, before the hand arrives in \textit{oven open} and after the hand releases it in \textit{oven close}, where it ends more than $80^\circ$ from the observed pose.
With both, the part moves ahead of the finger on frames that carry 41.0\% and 33.6\% of the measured force (\cref{tab:supp_contact_loss}).
A constant law thus misses the latch and cannot hold the oven door against gravity whether its parameters come from appearance or from the measured force.

\section{The Learned Law Across Simulators}
\label{sec:supp_sim2sim}

\noindent\textbf{Setup.}
For the closing plans of \cref{sec:exp_sim2sim}, the arm applies the planned force to each part at the contact point of its interaction, along the face normal for the hinged parts, at lever arms of 35.8\,cm, 22.2\,cm, and 22.7\,cm, and along the joint axis for the drawer.
We measure the joint coordinate from the first frame of the interaction in the closing direction.
The closed position is the closed stop of the twin, at $58.1^\circ$ for the microwave, $130.6^\circ$ for the laptop, and 47.4\,cm for the drawer, and for the oven the joint angle at the end of the interaction, $73.3^\circ$.
A part counts as closed if it reaches this position and stays within $1^\circ$ (1\,cm) of it or beyond it until the end of the 1\,s without force.

The simulated Franka arm runs in Isaac Lab~\cite{mittal2025isaaclab} with Newton~\cite{newton2025} at a physics step of $1/600$\,s.
Its Cartesian impedance controller keeps the end effector on the part at the contact point while the arm adds the planned force.
The arm executes every plan twice, once on the twin (Ours) and once on a standard joint in Newton with our reconstructed kinematics, joint limits, and learned resistance law.
On the twin, our simulator integrates \cref{eq:eom} in step-by-step co-simulation with the arm, and the two exchange the end-effector state and the contact force at every step.
In Newton, the MuJoCo solver simulates the joint and its contact with the end effector.

We repeat the comparison on the twins of the opening takes \textit{microwave open}, \textit{laptop open}, \textit{oven open}, and \textit{drawer open}, where the arm opens the part with a grasping tool.
We again plan with $w_f \in \{0.1, 1, 10\}$ on each twin and execute every plan on the twin (Ours) and on the standard joint in Newton.

\noindent\textbf{CEM planner.}
A plan is a force profile $F_{\text{plan}}(t)$ at the contact point, given by nine values at equal intervals over the 4\,s with linear interpolation.
We optimize these values by CEM on rollouts of \cref{eq:eom}, in which the arm is a point mass at the end effector with the inertia of the arm along the planned force.
The cost sums four terms, each averaged over its frames: the squared error to the closed position in the 1\,s without force and the squared coordinate past the closed position, both in degrees (cm for the drawer), the squared error to the closed position during the 4\,s of the plan in units of the stroke, and $w_f$ times the squared force in units of 10\,N.
Every plan also has to close the part with 90\% of its force.
A larger $w_f$ gives a slower plan: on the laptop, the peak planned force falls from 20\,N to 7\,N and the closing time rises from 1.2\,s to 3.3\,s, while the peak contact force stays at 2.8--3.0\,N (\cref{tab:supp_sim2sim}).
The microwave door and the laptop lid have no stop at the open position, the joint coordinate at the end of the interaction, and the oven door falls open under its own weight past $50^\circ$.
The arm therefore has to slow the part down and stop it at the open position, which it can only do with a grasp.
An opening plan is thus the trajectory of the grasped contact point to the open position, given by nine values over the 4\,s as for closing.
The cost has the same four terms, and $w_f$ now weighs the force that the arm applies along this trajectory.

\begin{table}[t]
  \caption{\textbf{Per-plan results of \cref{tab:sim2sim}.} Each plan (force penalty weight $w_f$) is executed on the twin in our simulator (Ours) and on a Newton~\cite{newton2025} joint with its learned law, and all 12 plans close the part in both simulators. We report the peak contact force in both simulators, the AAE and the largest difference of the joint coordinate ($^\circ$, cm for the drawer), and the RMSE of the generalized force (N$\cdot$m, N for the drawer) between them. Both force columns cover the frames before the part closes.}
  \vspace{-5pt}
  \centering
  \footnotesize
  \setlength{\tabcolsep}{0pt}
  \renewcommand{\arraystretch}{0.92}
  \begin{tabular*}{\linewidth}{@{\extracolsep{\fill}}l c cc cc c}
    \toprule
    & & \multicolumn{2}{c}{Peak force (N)} & \multicolumn{2}{c}{Joint coord.\ diff.} & Force diff. \\
    \cmidrule{3-4}\cmidrule{5-6}\cmidrule{7-7}
    Object & $w_f$ & Ours & Newton & AAE$\downarrow$ & Max$\downarrow$ & RMSE$\downarrow$ \\
    \midrule
    Microwave & 10  & 6.23 & 6.09 & 0.07 & 0.20 & 0.03 \\
              & 1   & 6.26 & 6.14 & 0.10 & 0.29 & 0.03 \\
              & 0.1 & 5.94 & 5.68 & 0.02 & 0.12 & 0.05 \\
    \midrule
    Laptop    & 10  & 3.02 & 3.03 & 0.27 & 1.62 & 0.01 \\
              & 1   & 2.91 & 2.91 & 0.48 & 3.82 & 0.01 \\
              & 0.1 & 2.77 & 2.71 & 0.18 & 2.05 & 0.02 \\
    \midrule
    Oven      & 10  & 3.12 & 3.12 & 0.46 & 2.06 & 0.03 \\
              & 1   & 3.10 & 3.10 & 1.26 & 3.38 & 0.04 \\
              & 0.1 & 3.11 & 3.11 & 0.60 & 1.36 & 0.09 \\
    \midrule
    Drawer    & 10  & 6.70 & 6.76 & 0.06 & 0.15 & 0.20 \\
              & 1   & 10.01 & 9.84 & 0.06 & 0.20 & 0.28 \\
              & 0.1 & 15.70 & 15.09 & 0.07 & 0.30 & 0.51 \\
    \bottomrule
  \end{tabular*}
  \label{tab:supp_sim2sim}
\end{table}

\noindent\textbf{Results.}
The transfer reported in \cref{sec:exp_sim2sim} holds for each of the twelve plans (\cref{tab:supp_sim2sim}).
The motion is the same up to a small time offset: the part reaches the closed position in the two simulators at most two frames of $1/30$\,s apart, which accounts for the largest differences of the joint coordinate, up to $3.8^\circ$ at 47--67$^\circ$/s.
Before the part closes, the force RMSE between them is 1--3\% of the largest generalized force of the plan on the microwave, the laptop, and the drawer and 4--13\% on the oven, and the peak forces in Newton stay within 5\% of those in our simulator, so the force-aware planning result of \cref{sec:exp_sim2sim} holds in Newton as well.
Without the arm, under the finger action of its take, the Newton joint follows the integration of \cref{eq:eom} with a mean difference of at most $0.03^\circ$ on the hinged parts and 0.17\,cm on the drawer, so a standard joint reproduces the learned law itself.

\begin{table}[t]
  \caption{\textbf{Cross-simulator comparison on the opening interactions.} Each opening plan (force penalty weight $w_f$) is executed in our simulator (Ours) and in Newton, and all 12 plans open the part in both simulators. We report the same quantities as in \cref{tab:supp_sim2sim}, and both force columns cover the frames before the part reaches the open position.}
  \label{tab:supp_open}
  \vspace{-5pt}
  \centering
  \footnotesize
  \setlength{\tabcolsep}{0pt}
  \renewcommand{\arraystretch}{0.92}
  \begin{tabular*}{\linewidth}{@{\extracolsep{\fill}}l c cc cc c}
    \toprule
    & & \multicolumn{2}{c}{Peak force (N)} & \multicolumn{2}{c}{Joint coord.\ diff.} & Force diff. \\
    \cmidrule{3-4}\cmidrule{5-6}\cmidrule{7-7}
    Object & $w_f$ & Ours & Newton & AAE$\downarrow$ & Max$\downarrow$ & RMSE$\downarrow$ \\
    \midrule
    Microwave & 10  & 9.92 & 9.93 & 0.00 & 0.07 & 0.05 \\
              & 1   & 10.02 & 10.04 & 0.00 & 0.05 & 0.05 \\
              & 0.1 & 10.69 & 10.69 & 0.00 & 0.00 & 0.01 \\
    \midrule
    Laptop    & 10  & 6.26 & 6.28 & 0.00 & 0.03 & 0.01 \\
              & 1   & 6.41 & 6.43 & 0.00 & 0.03 & 0.01 \\
              & 0.1 & 6.67 & 6.67 & 0.01 & 0.04 & 0.01 \\
    \midrule
    Oven      & 10  & 2.78 & 2.78 & 0.03 & 0.20 & 0.09 \\
              & 1   & 2.82 & 2.82 & 0.07 & 0.38 & 0.28 \\
              & 0.1 & 3.71 & 3.71 & 0.03 & 0.04 & 0.00 \\
    \midrule
    Drawer    & 10  & 16.97 & 18.11 & 0.02 & 0.17 & 0.69 \\
              & 1   & 22.07 & 22.79 & 0.02 & 0.14 & 0.55 \\
              & 0.1 & 29.68 & 30.23 & 0.01 & 0.11 & 0.55 \\
    \bottomrule
  \end{tabular*}
\end{table}

All twelve opening plans open the part and keep it open in both simulators, reaching the open position at most one frame apart, with peak forces within 0.02\,N on the hinged parts and 7\% on the drawer (\cref{tab:supp_open}).
Before the part reaches its end position, the force RMSE between them is at most 6\% of the largest generalized force of the plan, against at most 13\% for closing, except for the oven plan of $w_f = 1$ (18\%), which holds the door at rest for about 1\,s on its way.
Both simulators also agree on how each mechanism shapes the force: the microwave door leaves its latch at 9.9--10.7\,N at every $w_f$, the arm brakes the oven door with 5.7--5.8\,N past $50^\circ$, and the peak force on the drawer falls from 30\,N to 17--18\,N as $w_f$ rises from 0.1 to 10.
The laws learned from the opening takes thus act as simulator-independent joint models as well, also under braking, and plans optimized on these twins transfer to Newton.

\section{Generalization to an Unseen Contact Point}
\label{sec:supp_cosim}

\noindent\textbf{Setup.}
The CP1 interaction covers $58^\circ$ of door travel and the CP2 interaction the last $36^\circ$ of it, and we align the joint angles of the two at the closed stop, a shift of $22.0^\circ$.
In every test, the two twins share one joint model, with its inertia, gravity, joint limits, and lever arm, and differ only in the learned law, so each difference between them traces back to the law.
This joint model is that of the replayed interaction in the human replay and that of the CP2 interaction under the robot plans.
Over the first $22^\circ$ of the replay at CP1, which the CP2 interaction does not cover, the law of $\mathcal{T}_2$ keeps the friction at the end of its range and applies no conservative torque.
The simulated arm, its controller, and the step-by-step co-simulation are those of \cref{sec:supp_sim2sim}, with the planned force along the face normal of the door at CP1 or CP2, and we compare the force before the door reaches the stop.

\noindent\textbf{CEM planner.}
The CEM planner of \cref{sec:supp_sim2sim} optimizes a force profile $F_{\text{plan}}(t)$ over 4\,s on one twin, and its cost is the angle error to the closed stop, at $36.5^\circ$, at the end of the episode plus $w_f = 1$ times the squared force.

\noindent\textbf{Results.}
Under the robot plans of \cref{tab:cosim}, the two twins differ by 0.09--0.44$^\circ$ in door angle and by 0.15--0.20\,N$\cdot$m in generalized force, the level at which they reproduce the measured interactions in the human replay (0.11--0.86$^\circ$ and 0.06--0.23\,N$\cdot$m), so either twin can plan at both contact points.
The lowest force correlation of \cref{tab:cosim}, $r = 0.84$, comes from a time offset at the latch.
At CP2, the plan optimized on $\mathcal{T}_1$ carries the door through the latch at about $55^\circ$/s, and the door of $\mathcal{T}_2$ reaches the latch 27\,ms later, while the two latch peaks differ by 10\%.
At the latch, the peak of $\mathcal{T}_2$ is 10--15\% higher than that of $\mathcal{T}_1$ at both contact points, in line with the measured peaks of the two interactions, 2.40 and 2.78\,N$\cdot$m, which the twins reproduce with 2.37 and 2.69\,N$\cdot$m.
Each law thus carries the latch torque of its own interaction to the unseen contact point, where the lever arm sets the contact force, which confirms that the learned joint dynamics are independent of the contact point.

\section{Practical Cost}
\label{sec:supp_cost}

\begin{table}[t]
  \caption{\textbf{Practical cost.} Wall-clock time of the full pipeline on one interaction per object, from the recorded streams to the twin with its learned law, on one NVIDIA RTX 5090. Times in minutes, with the resistance law trained for 1000 steps on every object.}
  \centering
  \footnotesize
  \setlength{\tabcolsep}{3pt}
  \begin{tabular}{l cccc c}
    \toprule
    & Microwave & Laptop & Oven & Drawer & \\
    & close 2 & open & close & close & Mean \\
    \midrule
    Interaction (s) & 10.5 & 7.8 & 5.8 & 8.0 & 8.0 \\
    Cameras & 3 & 3 & 3 & 1 & \\
    \midrule
    Preprocessing & 27.0 & 14.7 & 13.7 & 6.4 & 15.4 \\
    \quad masks and point tracks & 21.3 & 10.2 & 10.2 & 5.0 & 11.6 \\
    \quad joint and part labels & 3.3 & 1.9 & 1.9 & 0.5 & 1.9 \\
    \quad hand fit & 2.4 & 2.6 & 1.6 & 0.9 & 1.9 \\
    Initial twin & 42.4 & 19.6 & 23.6 & 32.9 & 29.6 \\
    \quad generated meshes & 30.6 & 9.3 & 13.8 & 25.0 & 19.6 \\
    \quad Gaussians and particles & 11.8 & 10.4 & 9.8 & 8.0 & 10.0 \\
    Resistance law & 27.0 & 26.4 & 26.3 & 23.4 & 25.8 \\
    \midrule
    Total & 96.4 & 60.7 & 63.6 & 62.8 & 70.9 \\
    Peak GPU memory (GB) & 24.4 & 19.3 & 19.8 & 22.2 & 21.4 \\
    \bottomrule
  \end{tabular}
  \label{tab:supp_cost}
\end{table}

\Cref{tab:supp_cost} lists the cost of \cref{sec:exp_cost} per object and per stage, timed on a workstation with one NVIDIA RTX 5090 and an AMD Ryzen 9 9950X that runs one job at a time.
We calibrate the cameras once per session with a ChArUco board, and for each interaction the user provides a text prompt of the object for the segmentation, the joint type, the finger that carries the sensor, and the frame of each object state for the generated meshes.
The reconstruction takes 71\,min on average, between 61 and 96\,min, and at most 24\,GB of GPU memory.
Preprocessing grows with the length of the interaction and the number of cameras, from 6\,min for the drawer with one camera to 27\,min for the longest interaction.
The initial twin depends on the generated meshes, which take 9 to 14\,min for one state of the object and 25 to 31\,min when we generate the open and the closed state and compose them.
We time the resistance law at 1000 steps on every object, where it takes 23 to 27\,min.
For the action of one frame at 30\,Hz, our particle simulator returns the motion and the contact force of the microwave twin in 42\,ms, with 19{,}283 particles at a step of 0.21\,ms, and rendering the 0.5 to 1.7 million Gaussians of a twin at $848 \times 480$ pixels takes 2 to 5\,ms more.
A single rollout of \cref{eq:eom} takes 4\,ms per frame on the CPU, and 1024 rollouts of 5\,s run as one batch in 1.6\,s.
One plan of the CEM planner in \cref{sec:supp_sim2sim} evaluates 40{,}960 such rollouts, 40 iterations of 512 samples at two force scales, in 63 to 91\,s.

\end{document}